\pdfoutput=1
\documentclass[10pt,twocolumn,letterpaper]{article}

\usepackage[pagenumbers]{cvpr}

\usepackage{xcolor}
\definecolor{eccvblue}{rgb}{0.12,0.49,0.85}
\definecolor{cvprblue}{rgb}{0.21,0.49,0.74}

\usepackage{graphicx}
\graphicspath{{fig/}}

\usepackage{amsmath,amssymb,amsthm}
\usepackage{enumitem}
\newtheorem{proposition}{Proposition}

\theoremstyle{definition}
\newtheorem{definition}{Definition}
\newtheorem{assumption}{Assumption}
\theoremstyle{remark}
\newtheorem{remark}{Remark}

\usepackage{bm}
\usepackage{booktabs}
\usepackage{multirow}
\usepackage{array}
\usepackage{ulem}
\usepackage{pifont}
\usepackage{adjustbox}
\usepackage{placeins}

\newcolumntype{P}[1]{>{\centering\arraybackslash}p{#1}}

\usepackage{eccvabbrv}

\usepackage[pagebackref,breaklinks,colorlinks,citecolor=eccvblue,urlcolor=eccvblue]{hyperref}
\usepackage[capitalize]{cleveref}
\crefname{section}{Sec.}{Secs.}
\Crefname{section}{Section}{Sections}
\crefname{table}{Tab.}{Tabs.}
\Crefname{table}{Table}{Tables}
\crefname{figure}{Fig.}{Figs.}
\crefname{equation}{Eq.}{Eqs.}
\crefname{proposition}{Prop.}{Props.}
\crefname{corollary}{Cor.}{Cors.}
\crefname{assumption}{Assumption}{Assumptions}
\crefname{definition}{Def.}{Defs.}

\def\confName{CVPR}
\def\confYear{2027}

\begin{document}

\title{Forget the Data, Keep the Capability:\\
MoME, Switchable Unlearning for Retrieval}

\author{%
\begin{tabular}{@{}>{\centering\arraybackslash}p{\dimexpr\textwidth-4\tabcolsep\relax}@{}}
\mbox{Jingong Chen\textsuperscript{1,2}},
\mbox{Kyle Gao\textsuperscript{2}},
\mbox{Guangyuan Li\textsuperscript{4,2}},
\mbox{Chulwoo Pack\textsuperscript{1}},
\mbox{Sijia Liu\textsuperscript{3}},
\mbox{Kwanghee Won\textsuperscript{1,*}}\\[4pt]
{\normalsize
\mbox{\textsuperscript{1}South Dakota State University} \quad
\mbox{\textsuperscript{2}Aalto University} \quad
\mbox{\textsuperscript{3}Michigan State University} \quad
\mbox{\textsuperscript{4}ELLIS Institute Finland}
}
\end{tabular}
}

\maketitle

\begin{abstract}
Privacy compliance requires re-identification models to selectively forget individuals while preserving their ability to recognize others. For this, we present \textbf{MoME}, a switchable framework that can adapt pre-trained backbones for privacy compliance, allowing the same model to switch between \textbf{Full} mode for standard inference and \textbf{Core} mode for unlearned inference. MoME splits one backbone into \textit{core parameters} that preserve retrieval and zero-initialized \textit{parasitic parameters} that carry the full-data knowledge, and training them under \textit{gradient isolation} so the core never receives a forget-set gradient.
Disabling the branch yields \textbf{Core}, which satisfies a request to delete a person's training data, and enabling it yields \textbf{Full}, which retains what the deleted data taught, so honoring the request is a flag flip rather than a retraining run.
We prove \textbf{Core} independent of the forget set up to an audited set of coupling channels and introduce the rank-based \textit{Forgetting-Rank} statistic to audit that independence in a trained model.
On Market-1501, VeRi-776 and RegDB, \textbf{Core} reads at a retrained reference under three membership attacks while retrieving above it, and the retrieval advantage carries to ResNet and CLIP backbones.
Code and weights are available at \url{https://github.com/Jingong-Chen/MoME}.
\end{abstract}

\section{Introduction}
\label{sec::Introduction}

Machine unlearning removes the influence of specific training samples so that a model behaves as if it had never seen them~\cite{cao2015towards,bourtoule2021machine}. Privacy regulations increasingly require this capability, allowing individuals to request deletion of their personal data~\cite{voigt2017eu}. This requirement is particularly challenging for person and vehicle re-identification, where models learn identity-specific representations from large-scale datasets~\cite{zheng2015scalable,he2021transreid}. Once deployed, these representations are embedded in the model parameters, making selective deletion without degrading recognition performance a fundamental challenge.

\begin{figure}[t]
    \centering
    \includegraphics[width=0.85\linewidth]{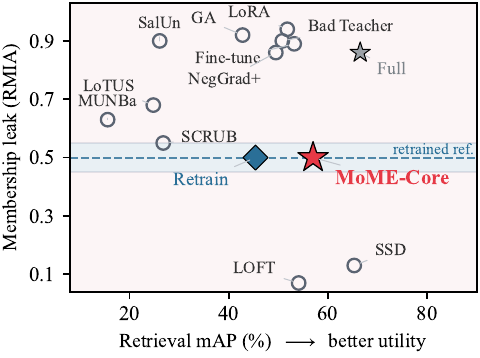}
\caption{\textbf{Other methods do not reach the utility and privacy frontier on ReID.} RegDB from visible to thermal, with each method placed by retrieval mAP and by membership leak under RMIA.
The dashed line is the retrained reference and the strip $\pm 0.05$ around it is what identity-clustered intervals resolve here.
Baselines are hollow circles and ours is the red star.}
    \label{fig::frontier}
\end{figure}

Existing methods remove influence only after training, by which point it is spread through the parameters and exact removal is intractable without retraining.
As \cref{fig::frontier} illustrates, the outcomes divide in three.
Methods that preserve retrieval either keep a strong membership signal, as a low-rank LoRA adapter~\cite{hu2022lora} and retain-set fine-tuning~\cite{warnecke2021machine} do, or drive it past the reference rather than to it, as SSD~\cite{foster2024fast} and LOFT~\cite{fang2026loft} do, which leaves them as distinguishable as a signal equally far above.
Methods that move the signal toward the reference do so by collapsing retrieval, as LoTUS~\cite{spartalis2025lotus} and MUNBa~\cite{wu2025munba} do, and stop short of it even so.
Retraining from scratch reaches the reference by definition but discards everything the forgotten data taught.

\begin{figure}[t]
    \centering
    \includegraphics[width=\linewidth]{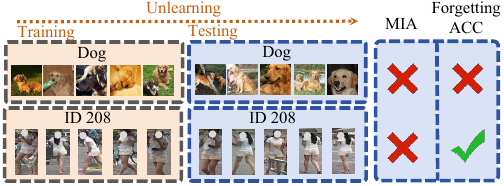}
\caption{\textbf{Forgetting the data and losing the capability coincide in \textit{classification} but not in \textit{retrieval}.}
Each mark reports what the metric reads after the request is honored.
Forgetting ACC is accuracy on the forgotten class above and retrieval accuracy on the forgotten identity below.
MIA, a membership inference attack, fails on both rows, the dog class is gone from the classifier, and the deleted identity is still retrieved, as it is by a model retrained without it.
Performance on the forgotten data carries the forgetting signal in classification and carries none in retrieval.}
    \label{fig::motivation}
\end{figure}

Retrieval adds a further difficulty, which \cref{fig::motivation} makes concrete.
A deleted class in classification is one of the model's own output categories, so the request names a unit the model has and removal, while not free, has a target.
Re-identification is evaluated on identities the model never saw, so nothing in the output corresponds to the person being deleted, and a model retrained without person $208$ ranks another image of them first $96.8\%$ of the time.
What that person contributed is embedding geometry built from pairwise constraints against every other identity, and the same geometry is what retrieves the unseen identities the model is tested on.
Deletion there removes influence from a space the model has to keep using, and performance on the deleted data is not what says whether it happened.

The cause is structural.
A post-hoc method starts from weights that already carry the forget set, so it offers an approximation of removal rather than a guarantee.
Acting during training keeps the forget set's gradients out of the deployed parameters, so the dependence never forms and removal becomes provable up to paths outside the gradient.
Re-identification is a natural setting for this, since the party that trains the model is the data controller obliged to honor the request, so the deletion set is known before training begins.
It also settles the reference.
The deleted images contribute real accuracy, so a model that still performs like the one trained on everything has not given that contribution up, and a method is credible when it lands on the model retrained without the deleted data.

Our key observation is that the two models a deletion request puts in tension can share one backbone.
We propose \textbf{M}ixture \textbf{o}f \textbf{M}ode \textbf{E}xperts (MoME), which augments the normalization affines and the linear maps with \textit{parasitic parameters} that carry the full-data knowledge, and trains them under \textit{gradient isolation} so the \textit{core parameters} never receive a forget-set gradient.
Disabling the branch at inference yields \textbf{Core}, provably independent of the forget set up to an audited set of coupling channels and reading at a retrained reference under three membership attacks and the \textit{Forgetting-Rank} statistic, within what these benchmarks resolve.
Enabling it yields \textbf{Full}, which recovers what the deleted data taught, so honoring the request is a flag flip rather than a retraining run.
The mechanism is not specific to retrieval and carries to a ResNet-50 backbone, to CLIP, and to class deletion on CIFAR-10.

We have three main contributions as follows:
\begin{itemize}[leftmargin=*]
\item We formulate unlearning for retrieval, where the deleted target is not an output category, so removal is certified against genuine retrains rather than by performance on the forget set.

\item We propose MoME, which adapts existing pre-trained backbones for privacy compliance by isolating full-data knowledge in zero-initialized \textit{parasitic parameters} attached to affine LayerNorm and BatchNorm~\cite{ioffe2015batch} layers. Under \textit{gradient isolation}, the backbone switches between \textbf{Full} and \textbf{Core} modes, while \textbf{Core} remains independent of the forget set except through an audited set of coupling channels.

\item We introduce \textit{Forgetting-Rank} $\rho_{\mathcal U}$, a rank statistic whose null is $\tfrac{1}{2}$ under exchangeability, testing whether forgotten samples incur losses consistent with models that never trained on them. We use it with three membership attacks to show on Market-1501, VeRi-776 and RegDB that \textbf{Core} reads at a retrained reference while retrieving above it, and that \textbf{Full} recovers the full-data accuracy.
\end{itemize}

\section{Related Work}
\label{sec::RelatedWork}

\subsection{Machine Unlearning}

Approaches divide by how much they guarantee~\cite{bourtoule2021machine,cao2015towards,nguyen2025survey}.
Exact methods partition data into independent shards~\cite{bourtoule2021machine,ginart2019making} but scale poorly to modern architectures, and certified removal bounds the influence that may remain~\cite{guo2020certified,sekhari2021remember}.
Approximate methods act on a trained model and include fine-tuning on the retain set~\cite{warnecke2021machine}, gradient ascent on the forget set~\cite{thudi2022unrolling}, random relabeling~\cite{graves2021amnesiac}, corrections from influence functions~\cite{koh2017understanding}, distillation from a teacher that diverges on the forget set~\cite{chundawat2023bad,kurmanji2023towards}, saliency-guided weight selection~\cite{fan2024salun}, dampening the synapses a sample relies on~\cite{foster2024fast}, instance-wise unlearning on pre-trained classifiers~\cite{cha2024learning}, sparsity~\cite{jia2023model}, and Fisher forgetting~\cite{golatkar2020eternal}.
The recent state of the art follows the same pattern, with LoTUS~\cite{spartalis2025lotus} calibrating output uncertainty to an information-theoretic bound, MUNBa~\cite{wu2025munba} resolving the forget and retain gradient conflict by Nash bargaining, and LOFT~\cite{fang2026loft} projecting features onto a low dimensional subspace.

All of them are built and scored on classification, where the forgotten target is one of the model's output categories.
The protocol that has settled around that setting reads accuracy on the forget set, accuracy on the retain and test sets, and a membership attack together~\cite{jia2023model,fan2024salun,kurmanji2023towards}, and the first of those is at once the forgetting signal and a loss of utility.
Retrieval does not admit that pairing, so the same suite cannot score it and a certificate there rests on the membership axis alone.

A concurrent work, MUNKEY~\cite{laguna2026munkey} also unlearns by design, routing memorization into a per-instance key so that deleting the key forgets that instance with no weight update.
Its unit is one training instance, its evaluation is classification, and the weights left after a key is deleted carry no independence statement, which is what a deletion request asks of a retrieval model whose deployed parameters are what ships.

\subsection{Re-Identification and Identity Deletion}

Re-identification retrieves a person or vehicle across non-overlapping cameras and has evolved from metric learning with CNNs~\cite{zheng2015scalable,liu2016deepveri} through part-based models and strong CNN baselines~\cite{sun2018beyond,luo2019bag,ye2021deep} to Vision Transformers such as TransReID~\cite{he2021transreid}, with cross-modal extensions~\cite{ye2018hierarchical,nguyen2017person,li2024day,wu2025relation} requiring invariance between visible and infrared cameras.
The privacy of the identities such a model memorizes remains underexplored.

Two efforts come close.
De-ReID~\cite{peng2025deid} solves inference-time identity rejection, where a variation-guided objective shifts the retrieval behavior of a fixed trained model so that designated persons are no longer matched, but the target identities stay in the training data, the weights keep the memory derived from them, and it is neither aimed at nor evaluated against membership inference.
TIU-ReID~\cite{chen2026tiureid} removes a target identity with a removable memory adapter and, like MoME, exposes a removal mode obtained by deleting a component at inference, but its adapter is fitted to a backbone that already trained on that identity, so the weights left after removal have seen it.
The three positions differ in whether the weights that ship ever saw the forget set, and only MoME's never did.

\section{Method}
\label{sec::Method}

MoME consists of three components. \textit{Parasitic parameters} turn a single backbone into two switchable modes, while \textit{gradient isolation} ensures that \textbf{Core} receives no gradient from the forget set, providing the basis for the unlearning guarantee. A consolidation stage then recovers the \textbf{Full} payload from the parameter difference relative to the original model. Finally, \textit{Forgetting-Rank} $\rho_{\mathcal U}$ audits the resulting forgetting behavior on the trained model.

\begin{figure*}[t]
 \centering
 \includegraphics[width=1.0\linewidth]{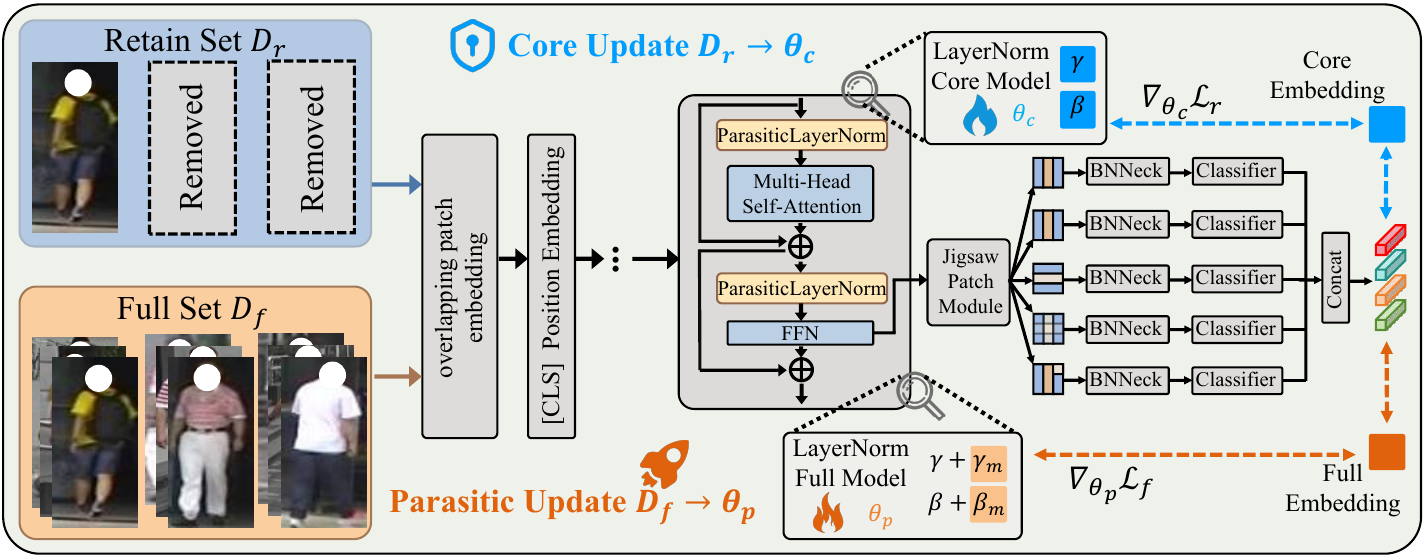}
\caption{Overview of the MoME training framework.
Each iteration runs two flows through one shared backbone: Step~1 forwards the retain set $\mathcal{D}_r$ in Core mode, with the \textit{parasitic parameters} $\boldsymbol{\theta}_p$ frozen, and updates the \textit{core parameters} $\boldsymbol{\theta}_c$; Step~2 forwards the full set $\mathcal{D}_f$ in Full mode, with $\boldsymbol{\theta}_c$ frozen, and updates $\boldsymbol{\theta}_p$; the diagram draws the normalization shadows, and the branch also carries a zero-initialized low-rank pair beside every backbone attention and FFN projection; a single optimizer step applies both updates.
A consolidation stage then initializes the parasitic payload at the parameter difference toward the original model and refines it, so toggling the branch on serves the Full mode and toggling it off yields Core, the parameters the guarantee is about.}
 \label{fig::framework}
\end{figure*}

\subsection{Preliminaries}
\label{sec::Preliminaries}

\noindent \textbf{Problem formulation.} Let $\mathcal{D}_f$ be the full training set of a person or vehicle re-identification model.
A deletion request names a set of training identities rather than a category the model predicts, so what has to go is their contribution to the weights and not the model's ability to match people.
The \textit{forget set} $\mathcal{U}$ holds every image of those identities, none of them survives in the \textit{retain set} $\mathcal{D}_r$, and the controller keeps only $k$ images per retained identity as a prototype set for continued service.
The three sets do not partition $\mathcal{D}_f$.
The images of retained identities beyond the $k$ kept, $4972$ of $12936$ on Market-1501, are named by no request and trained on by neither Core nor the reference, and membership is evaluated on $\mathcal{U}$ alone.
This matches how deletion arrives in re-identification, where a person revokes consent and all of their images must go, and the thin retain set is where retraining costs most.
The retrained reference trains from scratch on $\mathcal{D}_r$ alone and so never observed $\mathcal{U}$.
The target is a model whose membership behavior on $\mathcal{U}$ is indistinguishable from it at competitive retrieval accuracy.

\noindent \textbf{Baseline architecture.} We adopt TransReID~\cite{he2021transreid} on a ViT-B/16 backbone~\cite{dosovitskiy2020image} at stride $12$, whose five branches give a $3840$-dimensional descriptor at inference.

\subsection{Parasitic Parameter Mechanism}
\label{sec::Parasitic}

Retraining from scratch on $\mathcal{D}_r$ guarantees privacy but discards what the full dataset taught.
We observe that additive parameters attached to existing layers can be excluded from the forward pass exactly, which turns one backbone into a switchable pair of models.
This motivates \textit{parasitic parameters}, additive terms injected into the backbone and toggled at inference, as \cref{fig::framework} illustrates.
During training the branch hosts the Full mode of the two step procedure below; a consolidation stage then refits the slot toward the original model.

Formally, standard Layer Normalization~\cite{ba2016layer} computes
\begin{equation}
 \hat{\mathbf{x}} = \frac{\mathbf{x} - \mu}{\sqrt{\sigma^2 + \epsilon}}, \quad
 \mathbf{y} = \boldsymbol{\gamma} \odot \hat{\mathbf{x}} + \boldsymbol{\beta},
 \label{eq::ln}
\end{equation}
where $\boldsymbol{\gamma}$ and $\boldsymbol{\beta}$ are learnable affine parameters.
We augment each Layer Normalization with a pair of \textit{parasitic parameters} $\boldsymbol{\gamma}_m$ and $\boldsymbol{\beta}_m$, both initialized to zero so that the initial model behavior is unchanged.
The augmented forward pass becomes
\begin{equation}
 \mathbf{y} =
 \begin{cases}
 (\boldsymbol{\gamma} + \boldsymbol{\gamma}_m) \odot \hat{\mathbf{x}} + (\boldsymbol{\beta} + \boldsymbol{\beta}_m), & \text{Full mode,} \\
 \boldsymbol{\gamma} \odot \hat{\mathbf{x}} + \boldsymbol{\beta}, & \text{Core mode.}
 \end{cases}
 \label{eq::parasitic}
\end{equation}
Switching between the two modes is controlled by a single Boolean flag that enables or disables the parasitic branch across all augmented layers; in Core mode, the parasitic terms are excluded from the computation graph.

Within the backbone, the shadows attach to the LayerNorm layers preceding attention and the FFN in each Transformer block, as well as to the final normalization layer. Because LayerNorm affine parameters are diagonal, they only rescale the channels produced by the core and cannot mix them; we therefore extend the same additive principle to the linear projections. Each attention and FFN projection receives a zero-initialized low-rank branch following LoRA~\cite{hu2022lora}, which is active only in Full mode, giving $\mathbf{W}\mathbf{x}$ in Core mode and $\mathbf{W}\mathbf{x}+\frac{s}{r}\mathbf{B}\mathbf{A}\mathbf{x}$ in Full mode, where $s$ is the scale and $\mathbf{B}$ is initialized to zero. This construction leaves Core bit-identical to the unaugmented network while making Full identical to Core at initialization. Because the parasitic branches cannot affect Core and their rank determines the additional capacity available to Full, safety and capacity can be controlled independently.

\subsection{Gradient Isolation}
\label{sec::GradIso}

\textit{Parasitic parameters} provide the switchable parameterization but do not by themselves guarantee forgetting. If both modes are trained on the full dataset, the forget set can still update the shared backbone, leaving \textbf{Core} dependent on forgotten samples. The guarantee therefore comes from \textit{gradient isolation}, which restricts each data source to its designated parameter set.

Let $\boldsymbol{\theta}_c$ denote the core parameters and $\boldsymbol{\theta}_p$ the parasitic parameters. Each iteration consists of two forward-backward passes followed by a single optimizer step. In Step~1, we disable the parasitic branch, process a retain batch in \textbf{Core} mode with classification heads restricted to retained identities, and compute gradients only for $\boldsymbol{\theta}_c$. Thus, forgotten identities produce no updates to \textbf{Core}. In Step~2, we freeze $\boldsymbol{\theta}_c$, process a batch from $\mathcal{D}_f$ in \textbf{Full} mode, and compute gradients only for $\boldsymbol{\theta}_p$:

\begin{equation}
 \mathbf{g}_c = \nabla_{\boldsymbol{\theta}_c} \mathcal{L}_r,\;\;
 \nabla_{\boldsymbol{\theta}_p} \mathcal{L}_r = \mathbf{0};\quad
 \mathbf{g}_p = \nabla_{\boldsymbol{\theta}_p} \mathcal{L}_f,\;\;
 \nabla_{\boldsymbol{\theta}_c} \mathcal{L}_f = \mathbf{0}.
 \label{eq::step12}
\end{equation}
One step then applies both, $\boldsymbol{\theta}_c \leftarrow \boldsymbol{\theta}_c - \eta\,\mathbf{g}_c$ and $\boldsymbol{\theta}_p \leftarrow \boldsymbol{\theta}_p - \alpha\eta\,\mathbf{g}_p$.
Step~2 uses all of $\mathcal{D}_f$ rather than $\mathcal{U}$ alone, since the branch complements Core.
Every core gradient therefore comes from $\mathcal{D}_r$, and Core mode never evaluates $\boldsymbol{\theta}_p$.

\noindent \textbf{The Full payload.} The mechanism provides a separate additive parameter branch that is trained in two stages. During co-training, the branch is optimized for Step~2. Once the Core converges, a consolidation stage initializes each branch from its corresponding slice of
$\boldsymbol{\theta}_{\mathrm{orig}}-\boldsymbol{\theta}_c$, the parameter difference between the original model and the Core, and fine-tunes the branch on $\mathcal{D}_f$ while keeping the Core frozen. Batch Normalization statistics are the only state that is stored separately rather than represented as an additive parameter difference, since running estimates do not compose additively; the payload therefore stores the original statistics, and the mode switch selects between the two sets. This initialization is important because a zero-initialized branch must learn the entire parameter difference, whereas a branch initialized from that difference only needs to refine it. The Core remains bit-identical during consolidation because $\boldsymbol{\theta}_c$ is frozen throughout this stage. The resulting payload is dense because it stores the refined parameter difference without sparsification.

\noindent A gradient is not the only path by which data can influence model state. The backbone uses LayerNorm and maintains no running buffers, leaving the Batch Normalization layers in the classification heads as the only state outside the gradient path; Step~2 updates these statistics when the heads are left unfrozen. These statistics do not affect the Core representation: they carry no gradient, training normalizes using the current batch, and retrieval reads the feature before the BNNeck, so perturbing the running statistics leaves that feature bitwise unchanged. They do, however, affect the classifier logits exposed to the attacks, meaning that the attack surface is strictly larger than the released representation. For BatchNorm backbones, the running statistics can absorb information from $\mathcal{D}_f$, so freezing these statistics is necessary to close this additional information path.

\noindent Core is produced by a single end-to-end training stage initialized from ImageNet weights, without an adversarial objective or a separately trained student. After this stage, responding to the deletion request requires only disabling the Full branch. A later request targeting identities already encoded in the Core is a different problem: the mode switch cannot remove an influence already present in $\boldsymbol{\theta}_c$, so applying the isolation mechanism after training does not reach the required operating point and provides no corresponding guarantee. For privacy complience, only Core is released after deletion; the Full payload remains with the controller. Enabling the payload restores the removed capability, so the deletion guarantee applies to Core alone.

\noindent \textbf{Objective.}
\label{sec::Loss}
Both steps optimize the standard re-identification objective, a label-smoothed identity loss on each branch head~\cite{szegedy2016rethinking} together with a soft-margin triplet loss on the BNNeck features~\cite{luo2019bag,hermans2017defense}.
Step~1 adds the distillation term above.
Nothing in the objective is specific to unlearning: the separation between the two modes comes entirely from which parameters each step is allowed to move.

\begin{table*}[t]
\setlength{\tabcolsep}{4pt}
\renewcommand{\arraystretch}{0.9}
\setlength{\abovecaptionskip}{4pt}
\setlength{\belowcaptionskip}{0pt}
\centering
\caption{\textbf{Market-1501 and VeRi-776 under identity-level deletion with a thin retain set.} Half of the training identities are deleted, so $\mathcal{U}$ contains all images of the deleted identities, while each retained identity contributes $k{=}4$ images. Utility reports R-1 and mAP on the disjoint test set; privacy reports retrieval mAP on $\mathcal{U}$ and three attacks, two calibrated against the same retrained reference family. The reference row is evaluated against the other four reference models. \textbf{Full} denotes the consolidated parasitic branch. 
Every cell is a mean over five deletion splits with the spread across them, each split drawing its own half of the identities and contributing one run.
Attacks score each split against its own four retrained references, which share no checkpoint with the teachers; identity-level resampling within a split gives an interval of approximately $\pm0.023$. \textcolor{red}{Red}: Core; \textcolor{blue}{blue}: retrained reference; \textcolor{gray}{gray}: Full.}
\label{table::merged_market_veri}
\setlength{\tabcolsep}{0.7mm}
\scriptsize
\begin{tabular}{l cc cccc @{\hspace{3mm}} cc cccc}
\toprule
& \multicolumn{6}{c}{\textbf{Market-1501}} & \multicolumn{6}{c}{\textbf{VeRi-776}} \\
\cmidrule(lr){2-7}\cmidrule(lr){8-13}
& \multicolumn{2}{c}{Utility} & \multicolumn{4}{c}{Privacy} & \multicolumn{2}{c}{Utility} & \multicolumn{4}{c}{Privacy} \\
\cmidrule(lr){2-3}\cmidrule(lr){4-7}\cmidrule(lr){8-9}\cmidrule(lr){10-13}
Method & R-1 & mAP & $\mathcal{U}$ mAP & MIA$_\text{ent}$ & RMIA & U-LiRA & R-1 & mAP & $\mathcal{U}$ mAP & MIA$_\text{ent}$ & RMIA & U-LiRA \\
\midrule
Original & $94.8{\pm}0.0$ & $88.0{\pm}0.0$ & $99.9{\pm}0.0$ & $1.00{\pm}.00$ & $0.99{\pm}.00$ & $0.98{\pm}.00$ & $97.1{\pm}0.0$ & $82.3{\pm}0.0$ & $100.0{\pm}0.0$ & $1.00{\pm}.00$ & $0.89{\pm}.01$ & $0.81{\pm}.02$ \\
Fine-tune~\cite{warnecke2021machine} & $94.5{\pm}0.2$ & $86.9{\pm}0.2$ & $99.7{\pm}0.1$ & $0.98{\pm}.00$ & $0.98{\pm}.00$ & $0.97{\pm}.00$ & $91.8{\pm}0.7$ & $68.1{\pm}1.3$ & $94.8{\pm}1.2$ & $0.91{\pm}.02$ & $0.88{\pm}.03$ & $0.87{\pm}.03$ \\
GA~\cite{thudi2022unrolling} & $93.8{\pm}0.9$ & $85.2{\pm}1.6$ & $99.3{\pm}0.6$ & $0.91{\pm}.20$ & $0.90{\pm}.19$ & $0.88{\pm}.20$ & $39.9{\pm}29.7$ & $23.1{\pm}18.8$ & $25.8{\pm}22.0$ & $0.54{\pm}.08$ & $0.55{\pm}.06$ & $0.54{\pm}.05$ \\
NegGrad+~\cite{kurmanji2023towards} & $94.3{\pm}0.1$ & $86.3{\pm}0.2$ & $99.7{\pm}0.1$ & $1.00{\pm}.00$ & $0.99{\pm}.00$ & $0.97{\pm}.00$ & $34.0{\pm}7.5$ & $18.0{\pm}3.4$ & $13.6{\pm}3.0$ & $0.51{\pm}.04$ & $0.51{\pm}.03$ & $0.51{\pm}.03$ \\
SalUn~\cite{fan2024salun} & $92.8{\pm}0.3$ & $83.0{\pm}0.5$ & $99.0{\pm}0.3$ & $0.80{\pm}.03$ & $0.89{\pm}.01$ & $0.88{\pm}.01$ & $70.5{\pm}1.7$ & $42.9{\pm}1.7$ & $53.3{\pm}2.9$ & $0.60{\pm}.02$ & $0.60{\pm}.02$ & $0.60{\pm}.02$ \\
Bad Teacher~\cite{chundawat2023bad} & $92.3{\pm}0.3$ & $81.7{\pm}1.0$ & $98.8{\pm}0.4$ & $0.74{\pm}.03$ & $0.88{\pm}.01$ & $0.86{\pm}.01$ & $77.3{\pm}2.5$ & $50.2{\pm}2.1$ & $62.2{\pm}2.8$ & $0.52{\pm}.04$ & $0.51{\pm}.03$ & $0.51{\pm}.02$ \\
LoRA~\cite{hu2022lora} & $94.4{\pm}0.1$ & $87.0{\pm}0.3$ & $99.8{\pm}0.1$ & $1.00{\pm}.00$ & $0.99{\pm}.00$ & $0.98{\pm}.00$ & $39.9{\pm}33.0$ & $24.4{\pm}19.9$ & $25.5{\pm}23.8$ & $0.55{\pm}.12$ & $0.55{\pm}.09$ & $0.55{\pm}.08$ \\
SSD~\cite{foster2024fast} & $92.9{\pm}0.1$ & $83.2{\pm}0.2$ & $99.4{\pm}0.2$ & $0.22{\pm}.01$ & $0.28{\pm}.01$ & $0.31{\pm}.01$ & $8.6{\pm}2.0$ & $3.5{\pm}0.2$ & $2.6{\pm}0.4$ & $0.51{\pm}.02$ & $0.48{\pm}.01$ & $0.49{\pm}.01$ \\
SCRUB~\cite{kurmanji2023towards} & $94.1{\pm}0.0$ & $85.4{\pm}0.0$ & $99.7{\pm}0.1$ & $1.00{\pm}.00$ & $0.99{\pm}.00$ & $0.97{\pm}.00$ & $2.6{\pm}0.9$ & $1.5{\pm}0.7$ & $1.1{\pm}0.5$ & $0.50{\pm}.00$ & $0.49{\pm}.01$ & $0.50{\pm}.01$ \\
LoTUS~\cite{spartalis2025lotus} & $92.6{\pm}0.2$ & $81.1{\pm}0.3$ & $98.3{\pm}0.2$ & $0.56{\pm}.03$ & $0.80{\pm}.02$ & $0.75{\pm}.01$ & $10.3{\pm}0.6$ & $3.8{\pm}0.1$ & $3.0{\pm}0.1$ & $0.52{\pm}.01$ & $0.48{\pm}.01$ & $0.48{\pm}.01$ \\
MUNBa~\cite{wu2025munba} & $93.8{\pm}0.1$ & $84.7{\pm}0.1$ & $99.6{\pm}0.1$ & $0.97{\pm}.00$ & $0.98{\pm}.00$ & $0.97{\pm}.00$ & $44.8{\pm}2.0$ & $24.7{\pm}1.1$ & $23.6{\pm}1.0$ & $0.49{\pm}.02$ & $0.49{\pm}.01$ & $0.49{\pm}.01$ \\
LOFT~\cite{fang2026loft} & $94.3{\pm}0.2$ & $85.7{\pm}0.2$ & $99.6{\pm}0.1$ & $0.18{\pm}.01$ & $0.33{\pm}.01$ & $0.40{\pm}.01$ & $83.4{\pm}0.9$ & $41.6{\pm}2.8$ & $91.0{\pm}2.4$ & $0.32{\pm}.05$ & $0.47{\pm}.01$ & $0.48{\pm}.01$ \\
\midrule
\textcolor{blue}{Retrain} & \textcolor{blue}{$86.7{\pm}0.7$} & \textcolor{blue}{$71.4{\pm}0.9$} & \textcolor{blue}{$77.3{\pm}1.5$} & \textcolor{blue}{$0.48{\pm}.01$} & \textcolor{blue}{$0.50{\pm}.01$} & \textcolor{blue}{$0.50{\pm}.01$} & \textcolor{blue}{$65.0{\pm}6.0$} & \textcolor{blue}{$37.3{\pm}4.0$} & \textcolor{blue}{$32.8{\pm}4.7$} & \textcolor{blue}{$0.52{\pm}.02$} & \textcolor{blue}{$0.50{\pm}.02$} & \textcolor{blue}{$0.50{\pm}.01$} \\
\midrule
Core-only & $88.9{\pm}0.4$ & $75.2{\pm}0.7$ & $80.5{\pm}1.1$ & $0.49{\pm}.01$ & $0.51{\pm}.01$ & $0.51{\pm}.01$ & $69.5{\pm}1.0$ & $42.4{\pm}0.5$ & $38.1{\pm}1.4$ & $0.52{\pm}.01$ & $0.50{\pm}.02$ & $0.50{\pm}.02$ \\
\textbf{\textcolor{red}{MoME-Core}} & \textbf{\textcolor{red}{$89.0{\pm}0.5$}} & \textbf{\textcolor{red}{$75.3{\pm}0.7$}} & \textbf{\textcolor{red}{$80.5{\pm}1.1$}} & \textbf{\textcolor{red}{$0.49{\pm}.01$}} & \textbf{\textcolor{red}{$0.51{\pm}.01$}} & \textbf{\textcolor{red}{$0.52{\pm}.01$}} & \textbf{\textcolor{red}{$70.0{\pm}0.9$}} & \textbf{\textcolor{red}{$42.5{\pm}0.4$}} & \textbf{\textcolor{red}{$38.3{\pm}1.5$}} & \textbf{\textcolor{red}{$0.52{\pm}.01$}} & \textbf{\textcolor{red}{$0.50{\pm}.03$}} & \textbf{\textcolor{red}{$0.49{\pm}.02$}} \\
\textcolor{gray}{MoME-Full} & \textcolor{gray}{$95.2{\pm}0.1$} & \textcolor{gray}{$88.8{\pm}0.1$} & \textcolor{gray}{$99.8{\pm}0.1$} & \textcolor{gray}{$1.00{\pm}.00$} & \textcolor{gray}{$0.94{\pm}.00$} & \textcolor{gray}{$0.85{\pm}.00$} & \textcolor{gray}{$97.2{\pm}0.2$} & \textcolor{gray}{$81.4{\pm}0.1$} & \textcolor{gray}{$100.0{\pm}0.0$} & \textcolor{gray}{$1.00{\pm}.00$} & \textcolor{gray}{$0.87{\pm}.01$} & \textcolor{gray}{$0.78{\pm}.02$} \\
\bottomrule
\end{tabular}
\end{table*}

\subsection{Auditing the Guarantee: Forgetting-Rank}
\label{sec::rhodef}

\textit{Gradient isolation} guarantees forgetting by construction, but its effect must still be validated on the trained model. A reference-based attack tests whether \textbf{Core} is distinguishable from a retrained model, but this comparison conflates forgetting with generalization because co-training can yield lower loss on forgotten samples than vanilla retraining. We therefore introduce \textit{Forgetting-Rank}, a model-class-calibrated statistic that instead tests whether \textbf{Core} is exchangeable with models trained under the same procedure.

Let $\ell(x;\boldsymbol{\theta})$ denote the per-example cross-entropy, and let $\boldsymbol{\theta}^{(1)},\dots,\boldsymbol{\theta}^{(R)}$ denote shadow models trained using the same procedure. For each sample $x$, let $S(x)$ denote the shadows whose retain sets exclude $x$. On the retrieval benchmarks, all shadows share the same split, so $S(x)$ contains the full shadow set for every forgotten sample and exchangeability is evaluated over training randomness. On the classification benchmark, each shadow independently resamples the deleted classes, so $S(x)$ varies across samples and a sample is evaluated only when at least two shadows exclude it. For each forgotten sample, we rank the target model against the shadows that also exclude that sample,

\begin{equation}
\begin{split}
 \rho_{\mathcal U}
 = \frac{1}{|\mathcal U|}\sum_{x\in\mathcal U}\frac{1}{|S(x)|+1}&\Big[\tfrac12
 + \\ \sum_{j\in S(x)}\Big(\mathbf{1}\!\left[\ell_j(x)<\ell_c(x)\right]
 &+\tfrac12\mathbf{1}\!\left[\ell_j(x)=\ell_c(x)\right]\Big)\Big],
\end{split}
\label{eq::rho}
\end{equation}
the average normalized rank of the target's score, writing $\ell_c(x)=\ell(x;\boldsymbol{\theta}_c)$ and $\ell_j(x)=\ell(x;\boldsymbol{\theta}^{(j)})$.
Ties are assigned their mid-rank; for any $m$ exchangeable scores, the normalized mid-ranks sum to $m/2$, so their mean is $\tfrac12$ regardless of ties, whereas strict inequalities would assign tied scores an arbitrarily low rank. If \textbf{Core} is exchangeable with its shadows for every $x\in\mathcal U$, as guaranteed by \textit{gradient isolation}, each rank is uniform and $\mathbb{E}[\rho{\mathcal U}]=\tfrac12$ exactly, without distributional assumptions.

The statistic also detects controlled violations of forgetting. We train audit controls that retain a fraction of the requested identities during the Core update; the affected identities attain the lowest ranks, and ranking identities by mean rank places all fourteen identities retained under a $2\%$ violation among the top fifteen of $676$. In contrast, the aggregate $\rho_{\mathcal U}$ and calibrated attacks, evaluated on the same models, require a $5\%$ violation to detect the leakage. The controls use a $90\%$ deletion split and therefore characterize the sensitivity of the audit rather than the deployed operating point.

\begin{table*}[ht]
\setlength{\tabcolsep}{4pt}
\renewcommand{\arraystretch}{0.85}
\setlength{\abovecaptionskip}{4pt}
\setlength{\belowcaptionskip}{0pt}
\centering
\caption{\textbf{RegDB across modalities under identity-level deletion with a thin retain set}. Columns follow \cref{table::merged_market_veri}, with utility reported in both directions. The retrained reference uses AdamW, while the co-training recipe collapses on the $824$ retained images. 
Every cell is a mean over five deletion splits with the spread across them, as in \cref{table::merged_market_veri}, with attacks scored against each split's own four retrained references. RegDB contains $103$ deleted identities and $309$ identity clusters across the two pools, giving an attack-AUC resolution of approximately $\pm0.045$ when identities are the statistical unit.}
\label{table::comparison_regdb}
\setlength{\tabcolsep}{0.7mm}
\scriptsize
\begin{tabular}{l cc cc cccc}
\toprule
\multirow{2}{*}{Method} & \multicolumn{4}{c}{Utility} & \multicolumn{4}{c}{Privacy} \\
\cmidrule(lr){2-5}\cmidrule(lr){6-9}
 & V$\rightarrow$T R-1 & V$\rightarrow$T mAP & T$\rightarrow$V R-1 & T$\rightarrow$V mAP & $\mathcal{U}$ mAP & MIA$_\text{ent}$ & RMIA & U-LiRA \\
\midrule
Original & $67.1{\pm}0.0$ & $64.7{\pm}0.0$ & $68.3{\pm}0.0$ & $64.6{\pm}0.0$ & $100.0{\pm}0.0$ & $1.00{\pm}.00$ & $0.86{\pm}.01$ & $0.79{\pm}.01$ \\
Fine-tune~\cite{warnecke2021machine} & $54.9{\pm}2.2$ & $53.2{\pm}1.4$ & $52.1{\pm}2.7$ & $51.3{\pm}1.9$ & $100.0{\pm}0.0$ & $0.91{\pm}.01$ & $0.89{\pm}.02$ & $0.87{\pm}.02$ \\
GA~\cite{thudi2022unrolling} & $43.3{\pm}3.7$ & $42.8{\pm}3.0$ & $43.1{\pm}4.0$ & $42.4{\pm}3.2$ & $99.9{\pm}0.2$ & $1.00{\pm}.00$ & $0.92{\pm}.02$ & $0.88{\pm}.03$ \\
\midrule
NegGrad+~\cite{kurmanji2023towards} & $51.1{\pm}4.4$ & $49.5{\pm}4.1$ & $50.7{\pm}4.1$ & $50.4{\pm}4.3$ & $99.9{\pm}0.2$ & $1.00{\pm}.00$ & $0.86{\pm}.01$ & $0.81{\pm}.01$ \\
SalUn~\cite{fan2024salun} & $26.4{\pm}1.1$ & $26.1{\pm}1.5$ & $27.0{\pm}1.5$ & $27.4{\pm}1.5$ & $97.3{\pm}0.9$ & $0.93{\pm}.01$ & $0.90{\pm}.01$ & $0.87{\pm}.01$ \\
Bad Teacher~\cite{chundawat2023bad} & $53.1{\pm}2.6$ & $50.8{\pm}2.5$ & $54.5{\pm}3.7$ & $51.3{\pm}3.0$ & $100.0{\pm}0.1$ & $0.93{\pm}.03$ & $0.90{\pm}.01$ & $0.85{\pm}.01$ \\
LoRA~\cite{hu2022lora} & $53.3{\pm}8.7$ & $51.8{\pm}8.0$ & $53.1{\pm}9.3$ & $50.9{\pm}8.4$ & $100.0{\pm}0.0$ & $1.00{\pm}.00$ & $0.94{\pm}.06$ & $0.91{\pm}.07$ \\
SSD~\cite{foster2024fast} & $67.4{\pm}0.5$ & $65.3{\pm}0.4$ & $68.5{\pm}1.0$ & $64.9{\pm}0.5$ & $100.0{\pm}0.0$ & $0.06{\pm}.01$ & $0.13{\pm}.02$ & $0.15{\pm}.02$ \\
SCRUB~\cite{kurmanji2023towards} & $27.7{\pm}30.1$ & $26.8{\pm}27.6$ & $28.7{\pm}31.3$ & $28.0{\pm}29.1$ & $55.5{\pm}48.3$ & $0.55{\pm}.08$ & $0.55{\pm}.07$ & $0.53{\pm}.05$ \\
LoTUS~\cite{spartalis2025lotus} & $25.5{\pm}5.3$ & $24.8{\pm}5.3$ & $24.8{\pm}7.2$ & $24.1{\pm}5.5$ & $71.8{\pm}10.7$ & $0.66{\pm}.04$ & $0.68{\pm}.05$ & $0.64{\pm}.04$ \\
MUNBa~\cite{wu2025munba} & $16.4{\pm}1.7$ & $15.6{\pm}1.5$ & $16.3{\pm}1.4$ & $15.2{\pm}1.2$ & $59.2{\pm}7.5$ & $0.57{\pm}.05$ & $0.63{\pm}.04$ & $0.63{\pm}.03$ \\
LOFT~\cite{fang2026loft} & $56.1{\pm}1.6$ & $54.1{\pm}1.4$ & $57.2{\pm}1.6$ & $54.6{\pm}1.3$ & $100.0{\pm}0.0$ & $0.03{\pm}.01$ & $0.07{\pm}.01$ & $0.10{\pm}.01$ \\
\midrule
\textcolor{blue}{Retrain} & \textcolor{blue}{$47.0{\pm}3.8$} & \textcolor{blue}{$45.4{\pm}3.7$} & \textcolor{blue}{$47.8{\pm}4.1$} & \textcolor{blue}{$45.3{\pm}4.0$} & \textcolor{blue}{$51.8{\pm}3.6$} & \textcolor{blue}{$0.49{\pm}.02$} & \textcolor{blue}{$0.50{\pm}.01$} & \textcolor{blue}{$0.50{\pm}.02$} \\
\midrule
Core-only & $60.0{\pm}2.9$ & $57.7{\pm}2.3$ & $58.4{\pm}4.3$ & $55.8{\pm}3.2$ & $63.5{\pm}1.7$ & $0.50{\pm}.02$ & $0.50{\pm}.03$ & $0.50{\pm}.02$ \\
\textbf{\textcolor{red}{MoME-Core}} & \textbf{\textcolor{red}{$59.0{\pm}3.4$}} & \textbf{\textcolor{red}{$57.0{\pm}2.5$}} & \textbf{\textcolor{red}{$58.0{\pm}3.8$}} & \textbf{\textcolor{red}{$55.4{\pm}2.8$}} & \textbf{\textcolor{red}{$63.1{\pm}1.9$}} & \textbf{\textcolor{red}{$0.50{\pm}.01$}} & \textbf{\textcolor{red}{$0.50{\pm}.02$}} & \textbf{\textcolor{red}{$0.51{\pm}.01$}} \\
\textcolor{gray}{MoME-Full} & \textcolor{gray}{$68.1{\pm}0.3$} & \textcolor{gray}{$66.5{\pm}0.1$} & \textcolor{gray}{$70.4{\pm}0.6$} & \textcolor{gray}{$66.5{\pm}0.4$} & \textcolor{gray}{$100.0{\pm}0.0$} & \textcolor{gray}{$1.00{\pm}.00$} & \textcolor{gray}{$0.86{\pm}.01$} & \textcolor{gray}{$0.79{\pm}.01$} \\
\bottomrule
\end{tabular}
\end{table*}

\section{Experiments}
\label{sec::Experiments}

\subsection{Experimental Setup}
\noindent\textbf{Datasets.} We evaluate on \textbf{Market-1501}~\cite{zheng2015scalable}, \textbf{VeRi-776}~\cite{liu2016deepveri} and the cross modal \textbf{RegDB}~\cite{nguyen2017person}, with $751$, $576$ and $206$ identities, RegDB on the first of its ten standard splits.
A random half is deleted and each survivor keeps $k{=}4$ images, leaving $1491$ retained images against $6473$ forgotten on Market-1501, $1152$ against $18295$ on VeRi-776 and $824$ against $2060$ on RegDB.
The retain set is thinned on purpose.
An erasure request does not compel it: deleting one person's images says nothing about how much of everyone else's a controller keeps, and it is a stress regime where retraining is weakest.
The margin belongs to it: unthinned, in one run, the retrain is strong and Core only matches it, $82.5$ against $82.2$.

\noindent\textbf{Metrics.} Utility is scored using Rank-$k$ and mAP on the test set.
Retrieval on the deleted identities is reported as $\mathcal{U}$ mAP, read against the retrained reference rather than as a quantity to drive down.
Privacy is the ROC-AUC of three attacks on $\mathcal{U}$, entropy-MIA~\cite{shokri2017membership,salem2019mlleaks,song2021systematic}, RMIA~\cite{zarifzadeh2024low} and U-LiRA~\cite{carlini2022membership}.
Each of them is read together with its true positive rate at a fixed low false positive rate.
Non-members are test identities, whose classes the head does not contain, so each sample is scored against the model's own prediction rather than a class it has no column for.

\noindent\textbf{Implementation.} We follow the TransReID~\cite{he2021transreid} recipe on an ImageNet~\cite{deng2009imagenet} pre-trained ViT-B/16~\cite{dosovitskiy2020image}, with SGD on one NVIDIA RTX 5090: $120$ epochs, $40$ on VeRi-776, at batch $32$ and base learning rate $8\times10^{-3}$.
Remaining hyperparameters are in the configuration files.


\subsection{Comparison to Unlearning Baselines}
\label{sec::comparison}
We compare against \textbf{Original} without unlearning, \textbf{Fine-tune}~\cite{warnecke2021machine} on $\mathcal{D}_r$, gradient ascent \textbf{GA}~\cite{thudi2022unrolling} on $\mathcal{U}$, \textbf{Retrain} from scratch on $\mathcal{D}_r$, and nine recent unlearning methods. Neither Original nor Retrain provides a valid zero-leakage target, 
since Retrain still ranks another image of the deleted identities first in $90.3\%$ of Market-1501 queries. We therefore report each method relative to the Retrain reference rather than relative to zero or Original.

\cref{table::merged_market_veri} and \cref{table::comparison_regdb} reveal two baseline behaviors: methods that retain retrieval performance for the deleted identities at approximately the Original level, and methods that approach Retrain by reducing it.
Which behavior a method shows depends on the benchmark, not only on the method, and no method stays in the same group across all three.
On Market-1501 every baseline falls into the first group, the lowest being LoTUS at $98.3$ mAP on the deleted identities against $77.3$ for Retrain.
On VeRi-776 the same methods separate: SCRUB reaches $1.1$ mAP on VeRi-776, below the $32.8$ of Retrain, while on RegDB no baseline falls below the Retrain level at all.

\begin{figure}[t]
    \centering
    \includegraphics[width=\linewidth]{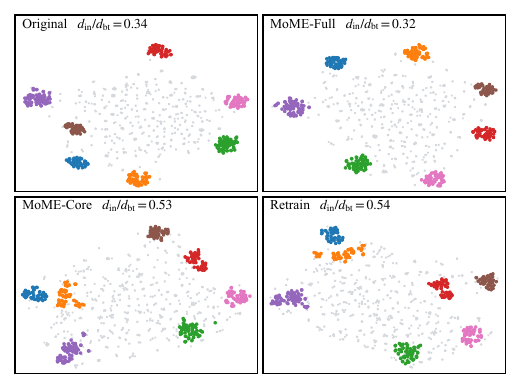}
\caption{\textbf{The branch moves the deleted identities between two geometries.} Seven deleted identities in color over a gray sample of retained images.}
    \label{fig::tsne}
\end{figure}

Core reads at the Retrain reference on every attack.
Over the three benchmarks, five deletion splits and three attacks the difference averages $+0.006$ and never exceeds $0.032$, and on VeRi-776 it falls on both sides of zero.
For scale, the methods that preserve retrieval on Market-1501 read between $0.98$ and $0.99$ on the same attack against $0.50$ for the reference.
Core achieves higher retrieval performance than the reference on all three datasets. 
The largest retrieval margin occurs on RegDB, where the thin retain set reduces standalone Retrain performance by $19.3$ mAP.
Enabling the Full branch restores the original retrieval performance, reaching $99.8$ mAP on the deleted identities against Original's $99.9$. \cref{fig::tsne} reads the same pairing off the embedding itself, as the mean distance of an image to its own identity center over the mean distance between identity centers, on all $376$ deleted identities of Market-1501. A spread alone would not separate these cases, since the centers contract as well.

The results also expose two limitations of evaluating unlearning with a single privacy metric. 
SSD and LOFT have calibrated attack scores of $0.28$ and $0.33$ on Market-1501, far below the null, while retaining $99.4$ and $99.6$ mAP on the deleted identities. Thus, the attack statistic can indicate successful removal even when retrieval performance remains close to Original. AUC also averages the entire operating curve, so we additionally examine the low-FPR regime: at a $1\%$ false-positive rate, RMIA recovers $1.0\%$ of deleted images from Core and $1.2\%$ from Retrain, compared with $78.4\%$ from Original.

Two close numerical values do not establish equivalence, and confidence intervals should not be estimated by resampling images independently because images from the same identity share a subject and camera set. We therefore resample identities and evaluate Core against the variability among five independent retraining runs, with each run ranked against the other four. Ten of the twelve cells place Core within $1.5$ standard deviations of the retraining distribution. The exception in the leakage direction is the true-positive rate on VeRi-776, where Core achieves $0.0156$ compared with $0.0096\pm0.0019$ for the retraining distribution; we therefore do not claim equivalence for this cell.

\begin{table}[t]
\renewcommand{\arraystretch}{0.85}
\setlength{\abovecaptionskip}{4pt}
\setlength{\belowcaptionskip}{0pt}
\centering
\caption{\textbf{Class deletion on image-classification benchmarks.} A tenth of the classes is deleted and $\mathcal{U}$ is every image of those classes.
Accuracy in \%, entropy-MIA of $\mathcal{U}$ against the test split, read as distance to the retrained reference.
MIA below $\tfrac12$ reflects a train and test difficulty gap, not stronger privacy. Single runs at seed $42$.}
\label{tab:classbench}
\setlength{\tabcolsep}{0.75mm}
\footnotesize
\begin{tabular}{l ccc ccc ccc}
\toprule
& \multicolumn{3}{c}{CIFAR-10} & \multicolumn{3}{c}{CIFAR-100} & \multicolumn{3}{c}{TinyImageNet} \\
Method & retain & $\mathcal{U}$ & MIA & retain & $\mathcal{U}$ & MIA & retain & $\mathcal{U}$ & MIA \\
\midrule
Original & 94.2 & 100.0 & 0.44 & 76.4 & 100.0 & 0.68 & 65.0 & 99.9 & 0.70 \\
\midrule
Fine-tune & 94.3 & 79.3 & 0.07 & 77.4 & 91.3 & 0.32 & 65.4 & 8.0 & 0.20 \\
GA & 11.1 & 0.0 & 0.50 & 1.1 & 0.0 & 0.50 & 0.6 & 0.0 & 0.50 \\
NegGrad+ & 12.7 & 0.0 & 0.50 & 2.5 & 0.0 & 0.53 & 0.6 & 0.0 & 0.47 \\
SalUn & 94.4 & 10.0 & 0.01 & 76.6 & 63.8 & 0.07 & 65.0 & 95.6 & 0.44 \\
Bad Teacher & 92.1 & 0.2 & 0.03 & 70.1 & 1.2 & 0.08 & 59.5 & 74.4 & 0.36 \\
SCRUB & 93.8 & 0.0 & 0.24 & 76.9 & 0.0 & 0.28 & 65.0 & 0.0 & 0.20 \\
\midrule
\textcolor{blue}{Retrain} & \textcolor{blue}{94.3} & \textcolor{blue}{0.0} & \textcolor{blue}{0.26} & \textcolor{blue}{77.7} & \textcolor{blue}{0.0} & \textcolor{blue}{0.24} & \textcolor{blue}{65.3} & \textcolor{blue}{0.0} & \textcolor{blue}{0.22} \\
\textcolor{red}{MoME-Core} & \textcolor{red}{94.1} & \textcolor{red}{0.0} & \textcolor{red}{0.31} & \textcolor{red}{77.3} & \textcolor{red}{0.0} & \textcolor{red}{0.24} & \textcolor{red}{65.3} & \textcolor{red}{0.0} & \textcolor{red}{0.22} \\
\bottomrule
\end{tabular}
\end{table}

\noindent Every competing method was designed for classification, so we reimplement all methods on a common ResNet-18 and evaluate class deletion at multiple scales on CIFAR-10, CIFAR-100~\cite{krizhevsky2009learning}, and TinyImageNet. No single metric separates the methods in \cref{tab:classbench}: gradient-ascent methods reach the null only by degrading the model, while SalUn falls below the retraining reference on the two smaller datasets. SCRUB is the exception, remaining near the reference at all three scales: on CIFAR-10 it achieves $0.24$ versus $0.26$ for Retrain, compared with $0.31$ for Core, while Core matches Retrain within $0.004$ on the other two datasets.

Attack AUC measures whether samples are distinguishable, whereas $\rho_{\mathcal U}$ in \cref{eq::rho} measures whether a model behaves like retraining under the requested deletion. We therefore report $\rho_{\mathcal U}$ only for Core and use the attacks for cross-method comparison. Core achieves $0.538$, $0.487$, and $0.494$, all within the range of their matched retraining families. Although the family mean is always $\tfrac12$ by construction, its spread provides the relevant reference. The audit controls confirm sensitivity: when the model retains $\mathcal{U}$, the readout falls to the bottom of the rank scale. This remains detectable for a small request: restricting the audit to five of $376$ deleted identities keeps the leaking readout at the floor, while the calibrated attacks no longer resolve a $0.3$ signal.

\begin{table*}[t]
\renewcommand{\arraystretch}{0.85}
\setlength{\abovecaptionskip}{4pt}
\setlength{\belowcaptionskip}{0pt}
\centering
\caption{\textbf{Generality across backbones, pretraining, and tasks.} MoME-Core against a matched reference retrained on $\mathcal{D}_r$. Utility is mAP for ReID and top-1 accuracy over all ten CIFAR-10 classes; the five-class deletion here is not comparable to the one-class setting in \cref{tab:classbench}. Privacy is $\rho_{\mathcal U}$ from \cref{eq::rho}, each column ranked against shadows from its own procedure. Five seeds on the deployed deletion split, which is why the ViT row differs from \cref{table::merged_market_veri}.}
\label{tab:generality}
\setlength{\tabcolsep}{5.2pt}
\footnotesize
\begin{tabular}{llll cc ccc}
\toprule
 & & & & \multicolumn{2}{c}{Utility} & \multicolumn{3}{c}{Privacy, $\rho_{\mathcal U}$} \\
\cmidrule(lr){5-6}\cmidrule(lr){7-9}
Backbone & Norm & Pretrain & Task & Core & Retr. & Core & Retr. & Full \\
\midrule
ViT-B/16          & LN & ImageNet    & ReID, Mkt  & 74.2 & 70.5 & $0.500 \pm .023$ & $0.500 \pm .085$ & $0.100 \pm .000$ \\
ResNet-50 AGW   & BN & ImageNet    & ReID, Mkt  & 47.5 & 33.3 & $0.500 \pm .254$ & $0.500 \pm .241$ & $0.880 \pm .015$ \\
CLIP ViT-B/16     & LN & contrastive & ReID, Mkt  & 77.7 & 67.6 & $0.500 \pm .128$ & $0.500 \pm .232$ & $0.896 \pm .005$ \\
ResNet-18         & BN & scratch     & CIFAR-10    & 47.8 & 47.8 & $0.502 \pm .056$ & $0.501 \pm .050$ & $0.152 \pm .005$ \\
\bottomrule
\end{tabular}
\end{table*}

\subsection{Ablation Study}
\label{sec::Ablation Study}

All ablations use Market-1501 on the deployed deletion split at seed $42$ and $100$ epochs, which is why the rows sit below \cref{table::merged_market_veri}, and the Step~2 control is repeated on RegDB; component ablations are in \cref{table::ablation_components}.

\begin{table}[t]
\setlength{\tabcolsep}{4pt}
\renewcommand{\arraystretch}{0.85}
\setlength{\abovecaptionskip}{4pt}
\setlength{\belowcaptionskip}{0pt}
\centering
\caption{\textbf{Ablation of necessary components} on Market-1501. MIA$_\text{ent}$ is the entropy attack on $\mathcal{U}$. Without Step~2 the branch stays at initialization, so Full equals Core and the row is a retrain on $\mathcal{D}_r$. Full is the co-trained branch before consolidation; the consolidated payload is in \cref{table::merged_market_veri}.}
\label{table::ablation_components}
\setlength{\tabcolsep}{1.6mm}
\footnotesize
\begin{tabular}{l ccc}
\toprule
Variant & Core mAP & Full mAP & MIA$_\text{ent}$ \\
\midrule
w/o Gradient Isolation & 85.1 & 88.3 & 0.995 \\
w/o Step 2 & 74.3 & 74.3 & 0.491 \\
Frozen backbone & 2.4 & 73.7 & 0.506 \\
\midrule
Parasitic $\gamma$ only & 75.4 & 78.1 & 0.487 \\
Parasitic $\beta$ only & 75.4 & 77.8 & 0.487 \\
\textbf{Ours} $\gamma{+}\beta$ co-trained & 74.4 & 81.5 & 0.493 \\ 
\bottomrule
\end{tabular}
\end{table}

\noindent \textbf{Gradient isolation.} Removing \textit{gradient isolation} raises Core to $85.1$ mAP, close to the $88.0$ of a model that never unlearned, and the attack reads $0.995$: without isolation the core trains on $\mathcal{D}_f$ too, so it does not forget.
The pair still reaches $88.3$ in Full, so the row does not fail for want of capacity.
It fails because the capacity sits in the core rather than in the branch, which is the one arrangement the switch cannot undo.
Isolation costs $10.7$ mAP of Core and is what moves the attack to the null.

\noindent \textbf{Step~2.} Removing it leaves Core where it was: the Core-only row of \cref{table::merged_market_veri} and \cref{table::comparison_regdb} sits within $1.0$ of Core on every retrieval column and within $0.011$ on every attack column.
Nor does it cost Full: consolidating from that core reaches $88.7$ mAP, the same as from the co-trained one.
The payload needs a core, not a second step, and the $3.8$ mAP over a plain retrain comes from how Step~1 trains on $\mathcal{D}_r$.
What Step~2 buys is the co-trained branch rather than the core or the payload: without it the shadows stay at their zero initialization, so Full equals Core at $74.3$, where co-training carries Full to $81.5$.

\noindent \textbf{A frozen backbone.} Freezing it collapses Core to $2.4$ while Full still reads $73.7$.
Step~1 then has nothing to adapt, so the core is the pre-trained initialization with heads on top and the $73.7$ is the branch carrying $\mathcal{U}$ on features it did not shape.
The attack still reads $0.506$, so safety does not rest on the core being any good: the core never sees $\mathcal{D}_f$ whether or not it can learn from what it does see.
What the row gives up is utility, and that is the part Step~1 takes from $\mathcal{D}_r$.

\noindent \textbf{One shadow or both.} A scale shadow alone and a shift shadow alone leave the same core, because Step~2 cannot reach it and Core mode excludes the branch, so both read $75.4$ mAP of Core and the two rows differ only in Full, at $78.1$ and $77.8$.
Which affine the branch is given therefore does not matter; how much of one it is given does.
Neither reaches the capacity of the pair: co-training raises Full to $81.5$ for $1.0$ mAP of Core.
All three sit at the null, so capacity is set by the parasitic budget and safety by whether the core ever sees $\mathcal{D}_f$.

\subsection{Generality across foundation model backbones}
\label{sec::generality}

MoME needs only affine LayerNorm or BatchNorm parameters, so the mechanism should transfer beyond the ImageNet pre-trained ViT.
We repeat the comparison on a ResNet-50 AGW backbone, on CLIP-ReID~\cite{radford2021learning,li2023clipreid} and on a ResNet-18 trained from scratch for CIFAR-10 class deletion, each against a reference retrained on its own $\mathcal{D}_r$.
\cref{tab:generality} reads the four settings.
Core matches or beats that reference on utility, and its \textit{Forgetting-Rank} sits at the null inside the band the reference family spans.
On the two non-ViT backbones that band is too wide to resolve anything, so what transfers there is the retrieval advantage and the mechanism running, not a privacy result.
A BatchNorm backbone needs its running statistics frozen in Step~2, and these runs did not freeze them.
The switch does not transfer as widely: on ResNet-50 and CLIP the payload no longer restores the deleted identities, and on CIFAR-10 it carries their loss profile at chance accuracy, so Full remains a result for the pre-trained ViT.

\section{Conclusion}
\label{sec::Conclusion}

We present MoME, a switchable framework for unlearning by design that can be adapted to existing backbones. We additionally introduce \textit{Forgetting-Rank}, a statistic that tests whether the loss on forgotten samples is statistically indistinguishable from that of models trained without them. MoME separates a backbone into \textit{core parameters}, which preserve retrieval, and \textit{parasitic parameters}, which isolate forget-set knowledge. Under \textit{gradient isolation}, the Core is provably independent of the forget set up to paths outside the gradient, while Forgetting-Rank provides a rank-based audit for trained models. 
Across Market-1501, VeRi-776, and RegDB, the deployed Core reads at the retrained reference on every attack while achieving higher retrieval performance, on ResNet and CLIP backbones as well.


\noindent \textbf{Acknowledgements.} This research was partially  supported in part by South Dakota State University’s High Performance Computing resources.

\FloatBarrier   

{\small
\bibliographystyle{ieeenat_fullname}
\bibliography{ref}
}

\maketitlesupplementary
\appendix

\setcounter{topnumber}{4}
\setcounter{bottomnumber}{3}
\setcounter{totalnumber}{6}
\setcounter{dbltopnumber}{4}
\renewcommand{\topfraction}{0.92}
\renewcommand{\bottomfraction}{0.85}
\renewcommand{\textfraction}{0.06}
\renewcommand{\floatpagefraction}{0.70}
\renewcommand{\dbltopfraction}{0.92}
\renewcommand{\dblfloatpagefraction}{0.70}

\section{Gradient Isolation Yields Exact Unlearning}
\label{sec::theory}

We now show that the two step schedule of \cref{sec::GradIso} is not merely a heuristic: it renders the deployed Core parameters \emph{provably independent} of the forgotten data, up to a small enumerable set of coupling channels that we state and audit.
The induction below is short, and its content is in the assumption set: withholding forget-set gradients from the core is not the same as making it independent of the forget set, since data reaches parameters by paths outside the gradient.
Two such paths exist in this implementation, the Batch Normalization buffers in the classification heads and the decision of whether to apply an update, and the second had to be changed in the code before A5 held of it.

\subsection{Setup and notation}
Let $\bm{\theta}=(\bm{\theta}_c,\bm{\theta}_p)$ collect the core and parasitic parameters, and let $\bm{b}$ denote all state that is not updated by gradients, for example the BatchNorm running statistics in the heads.
Training uses the full set $\mathcal{D}_f$, the retain set $\mathcal{D}_r\subseteq\mathcal{D}_f$ the core is updated on, and the forget set $\mathcal{U}\subseteq\mathcal{D}_f$ named by the request, with $\mathcal{U}\cap\mathcal{D}_r=\emptyset$.
Everything below uses that disjointness alone.
Under the deployed protocol the retain side is thinned, so $\mathcal{D}_f\setminus\mathcal{D}_r$ is strictly larger than $\mathcal{U}$: it also holds images of retained identities that $\mathcal{D}_r$ did not keep, which no request covers and which the membership results treat as a separate population.
A single iteration $t$ performs two forward and backward passes and one optimizer step:
\begin{itemize}[leftmargin=1.1em,itemsep=1pt,topsep=1pt]
\item \textbf{Step 1, Core.} With $\bm{\theta}_p$ excluded from the forward pass in Core mode and $\mathrm{requires\_grad}(\bm{\theta}_p){=}\textsc{False}$, draw a batch $B_r^t\subseteq\mathcal{D}_r$ and form $\bm{g}_c^t=\nabla_{\bm{\theta}_c}\mathcal{L}(B_r^t;\bm{\theta}_c^t)$.
\item \textbf{Step 2, Full.} With $\mathrm{requires\_grad}(\bm{\theta}_c){=}\textsc{False}$, draw $B_f^t\subseteq\mathcal{D}_f$ and form $\bm{g}_p^t=\nabla_{\bm{\theta}_p}\mathcal{L}(B_f^t;\bm{\theta}_c^t,\bm{\theta}_p^t)$.
\item \textbf{Update.} $\bm{\theta}_c^{t+1}=\mathcal{U}_c(\bm{\theta}_c^t,\bm{g}_c^t,\bm{m}_c^t)$ and $\bm{\theta}_p^{t+1}=\mathcal{U}_p(\bm{\theta}_p^t,\bm{g}_p^t,\bm{m}_p^t)$, with disjoint optimizer states $\bm{m}_c,\bm{m}_p$.
\end{itemize}
Write $\omega_r$ for all randomness in Step 1, namely retain batch order, augmentation, and dropout masks.
Step 1 also reads a fixed ensemble of teachers, each trained on $\mathcal{D}_r$ alone, which enter as inputs derived from $\mathcal{D}_r$.

\begin{definition}[Forget set data independence]
\label{def::indep}
A training procedure producing a model $\phi$ achieves \emph{exact unlearning} of $\mathcal{U}$ if $\phi$ is a deterministic function of $(\bm{\theta}^0,\mathcal{D}_r,\omega_r)$ alone; equivalently, conditional on $\bm{\theta}^0$ and $\mathcal{D}_r$, replacing $\mathcal{U}$ by any $\mathcal{U}'$ of the same size leaves the distribution of $\phi$ unchanged.
The conditioning is not cosmetic: $\mathcal{D}_r$ and $\mathcal{U}$ are two parts of one dataset and need not be independent of each other, so the statement is non-interference with respect to $\mathcal{U}$ once the retain set and the initialization are fixed, and the set sizes are public.

Such a $\phi$ is distributed exactly as the same procedure run with Step 2 removed, which is a training run that never reads $\mathcal{U}$.
It is not thereby distributed as an arbitrary retain-only algorithm.
Two procedures can both ignore $\mathcal{U}$ and still produce different parameters, different losses and different attack scores, so agreement with a particular retrained baseline is an empirical claim about that baseline and does not follow from this definition.
The guarantee is also relative to the training run it describes: it says nothing about information about $\mathcal{U}$ that the initialization already carries, which for an ImageNet or CLIP backbone is not something this procedure can remove.
\end{definition}

We make explicit the assumptions under which isolation is exact; each names a coupling channel that our audit of eleven channels tests in practice.
\begin{assumption}\label{ass::channels}\leavevmode
\begin{enumerate}[label=A\arabic*,leftmargin=2.2em,itemsep=1pt,topsep=1pt]
\item \textbf{Purity of the Core forward.} The Core forward excludes $\bm{\theta}_p$; hence $\bm{g}_c^t$ is independent of $\bm{\theta}_p$.
\item \textbf{Core freeze in Step 2.} $\mathrm{requires\_grad}(\bm{\theta}_c){=}\textsc{False}$ in Step 2, so autograd contributes $\bm{0}$ to the core gradient there.
\item \textbf{No stateful leakage.} Backbone normalization is LayerNorm, which is per-sample and keeps no running buffers.
The only buffered norms are the head BNNeck, whose affine terms belong to $\bm{\theta}_c$ while their running statistics are buffers.
Those buffers carry no gradient, and the training forward runs in train mode, where BatchNorm normalizes by the statistics of the batch in hand rather than by the running estimates, so no core gradient is a function of them whether or not Step 2 updates them.
Freezing them in Step 2 keeps the buffers themselves free of $\mathcal{D}_f$ as well, which matters for the classifier logits and not for $\bm{\theta}_c$.
\item \textbf{Step 1 randomness from the retain set.} $\omega_r$, comprising sampling, augmentation, and dropout, is drawn over $\mathcal{D}_r$.
The two steps share one pseudo-random stream, so $\omega_r$ can depend on how many forget batches have been drawn, a count the set sizes fix, and on no sample of $\mathcal{U}$.
\item \textbf{Disjoint optimizer state and update decision.} $\bm{m}_c$ and $\bm{m}_p$ are maintained per-parameter and never shared, and whether an update is applied is decided separately for the two groups.
A decision taken jointly would be a channel rather than a count, since whether a batch produces a non-finite gradient depends on that batch; a non-finite Step 2 gradient must therefore drop $\bm{g}_p^t$ alone and leave $\bm{g}_c^t$ applied.
Mixed-precision loss scaling is one such shared decision and is disabled under the bfloat16 autocast these runs use.
\end{enumerate}
\end{assumption}

\begin{proposition}[Gradient isolation $\Rightarrow$ data independence]
\label{prop::indep}
Under \cref{ass::channels}, and with an initialization $\bm{\theta}^0$ that is independent of $\mathcal{U}$, for every iteration $t$ the core state $(\bm{\theta}_c^t,\bm{m}_c^t)$ is a deterministic function of $(\bm{\theta}^0,\mathcal{D}_r,\omega_r)$; in particular it is independent of the forget set $\mathcal{U}$.
The hypothesis on $\bm{\theta}^0$ is what the induction starts from and cannot be dropped: a core that begins as a function of $\mathcal{U}$ stays one under any later restriction of its gradients.
Hence the deployed Core model $f(\cdot;\bm{\theta}_c^T)$ achieves exact unlearning of $\mathcal{U}$ in the sense of \cref{def::indep}.
\end{proposition}
\begin{proof}
By induction on $t$.
\emph{Base.} $(\bm{\theta}_c^0,\bm{m}_c^0)$ is the ImageNet initialization, independent of all training data.
\emph{Inductive step.} Assume $(\bm{\theta}_c^t,\bm{m}_c^t)$ depends only on $(\bm{\theta}^0,\mathcal{D}_r,\omega_r)$.
In Step 1, A1 gives $\bm{g}_c^t=\nabla_{\bm{\theta}_c}\mathcal{L}(B_r^t;\bm{\theta}_c^t)$ with $B_r^t\subseteq\mathcal{D}_r$ selected by $\omega_r$ under A4; thus $\bm{g}_c^t$ is a function of $(\bm{\theta}_c^t,\mathcal{D}_r,\omega_r)$.
In Step 2, A2 sets the core gradient to $\bm{0}$, so Step 2 alters neither $\bm{\theta}_c$ nor $\bm{m}_c$.
It may update the head Batch Normalization buffers, which are left free in the runs reported here, but A3 puts those outside the induction: the training forward normalizes by the batch in hand, so no core gradient is a function of them.
The update $\mathcal{U}_c$ therefore combines only quantities that are local to the core and free of $\mathcal{U}$, as ensured by A5, and $(\bm{\theta}_c^{t+1},\bm{m}_c^{t+1})$ remains a function of $(\bm{\theta}^0,\mathcal{D}_r,\omega_r)$.
The forget set $\mathcal{U}$ enters the computation only through $B_f^t$ in Step 2, which updates $\bm{\theta}_p$ and, where the freeze is off, the head buffers, neither of which the induction carries.
The claim follows for all $t\le T$.
\end{proof}

\begin{remark}[Privacy parity, and why Core can exceed the reference]
\label{rem::paradox}
Step 2 leaves the core unchanged, so the Core trajectory coincides with the same procedure run with Step 2 removed, a run that reads $\mathcal{D}_r$ alone: Core is a bona fide estimator trained on $\mathcal{D}_r$, and its membership advantage on $\mathcal{U}$ is that of a procedure that never read $\mathcal{U}$.
Whether it also matches the particular retrained baseline reported in the tables is measured there and is not implied here.
Any separability that remains, visible as Core sitting at the reference's leak level in \cref{fig::frontier}, is generalization from $\mathcal{D}_r$ rather than memorization of $\mathcal{U}$.
The same proposition forbids any trace of $\mathcal{U}$ in $\bm{\theta}_c$ at the gradient level, so Core's \emph{utility} edge over the reference cannot come from leaked information either.
It reflects optimization differences on the retain side, distillation from an ensemble retrained on $\mathcal{D}_r$ alone and the pacing of the core's passes over it, and disabling Step 2 leaves Core unchanged, so it is not a co-training effect.
The proposition pins privacy by construction while leaving utility unconstrained.
\end{remark}

\begin{remark}[Scope]
\label{rem::scope}
\cref{prop::indep} is an \emph{exact} statement at the parameter level, unlike approximate or certified unlearning, but conditional on \cref{ass::channels}.
It confers no differential privacy with respect to $\mathcal{D}_r$, which Core legitimately learns, and says nothing about the Full model, which encodes $\mathcal{U}$ in $\bm{\theta}_p$ by design.
The audit of eleven coupling channels and three membership attacks, read at low false positive rates, exists to test A3 and A5 in practice.
\end{remark}

\subsection{A calibrated test for residual memorization}
\label{sec::ranktest}
\cref{prop::indep} makes forgetting \emph{testable}, and \cref{sec::rhodef} introduced the instrument: reference-based attacks confound Core's superior generalization with leakage, cf.\ \cref{rem::paradox}, so $\rho_{\mathcal U}$ of \cref{eq::rho} instead ranks each forgotten sample's cross-entropy among shadow models trained by the same procedure, using only the shadows $S(x)$ whose retain set excludes $x$.
Write $r(x)$ for the inner normalized rank in \cref{eq::rho}, so that $\rho_{\mathcal U}$ is its average over $\mathcal{U}$.
Here we prove that this calibration is exact.

\begin{proposition}[Exact calibration of $\rho_{\mathcal U}$]
\label{prop::rank}
Under \cref{ass::channels}, for every $x\in\mathcal{U}$ the scores $\{\ell(x;\bm{\theta}_c)\}\cup\{\ell(x;\bm{\theta}^{(j)}):j\in S(x)\}$ are exchangeable, so the mid-rank $r(x)$ of \cref{eq::rho} has $\mathbb{E}[r(x)]=\tfrac12$ with $m=|S(x)|+1$, and is uniform on $\{\tfrac{1}{2m},\tfrac{3}{2m},\dots\}$ when the scores are almost surely distinct; hence $\mathbb{E}[\rho_{\mathcal U}]=\tfrac12$.
The calibration is exact per sample and free of any distributional assumption.
It does not by itself fix a level for a test on the average, because the terms of $\rho_{\mathcal U}$ are not independent: one model's losses are correlated across samples, and images of one identity are correlated with each other.
We therefore read $\rho_{\mathcal U}$ against the spread the reference family itself shows rather than against a nominal level.
\end{proposition}
\begin{proof}[Proof sketch]
By \cref{prop::indep} the target core parameters are a deterministic function of $(\bm{\theta}^0,\mathcal{D}_r,\omega_r)$ and independent of $\mathcal{U}$; for $x\in\mathcal{U}$ this holds for the target and for each shadow $j\in S(x)$, all of which excluded $x$ from their retain set.
On the retrieval benchmarks they share one retain set and differ only in training randomness, an i.i.d.\ draw; on the classification benchmark the retain sets are i.i.d.\ draws as well.
Either way the induced scores $\ell(x;\cdot)$ are i.i.d.\ conditional on what the family shares, and thus exchangeable; the rank of one exchangeable variable among $m$ is uniform, giving $\mathbb{E}[r(x)]=\tfrac12$.
Averaging and the exactness of rank tests under exchangeability yield the claim.
\end{proof}

Two properties make $\rho_{\mathcal U}$ the right instrument.
\emph{It is unconfounded:} the target and its calibrating shadows share architecture, objective and optimizer, so any effect of generalization or of optimization quality shifts their scores together and cancels in the rank; only genuine training on $x$, which \cref{prop::indep} forbids for Core, can bias $r(x)$ downward.
\emph{It is powerful:} reading the \emph{same} network in Full mode, whose parasitic branch \emph{was} trained on $\mathcal{U}$, drives $\rho_{\mathcal U}$ from $0.500$ to $0.100$, the floor of the statistic at five calibrating shadows, so on almost every forgotten sample the leaking readout is the most confident model in the comparison.
Taking the identity as the unit, which the images of one person require, every one of the $1880$ identity readings falls below the null and $1877$ sit at the floor.
We calibrate against the reference family's measured spread rather than against the analytic null, because the scores are only approximately exchangeable: that family's own spread across targets is $0.085$ on Market-1501, $5.8\times$ what exchangeability predicts with the identity as the unit, since training randomness varies the models more than the assumption allows.
Read against that spread the test resolves a move of about $0.17$, two of its standard deviations, and the leaking readout moves $\rho_{\mathcal U}$ by $0.40$.

\section{Additional Experiments and Analyses}
\label{sec::supp}

\noindent\textbf{Reading the tables in this appendix.} Six conventions govern the numbers below.
First, the Forgetting-Rank $\rho_{\mathcal U}$ is de-confounded and ideal at exactly $\tfrac12$, so for a single model ranked against shadows that never trained on the sample its quality is the distance $|\rho_{\mathcal U}-\tfrac12|$, with a value below $\tfrac12$ indicating residual memory of the forget set and a value above it indicating over-forgetting.
A family average is a different object, $\tfrac12$ by a counting identity whatever its members did, so where a table reports one it is the spread that carries the information and not the distance from $\tfrac12$.
Second, the calibrated attacks RMIA and U-LiRA are judged against a genuine retrain rather than against an absolute target: this is the retrained reference row, which sits near $\tfrac12$ in the re-identification tables once the loss-based attacks score each sample against the model's own prediction.
The uncalibrated entropy-MIA is read the same way, against the reference in its own table rather than against a fixed number, since its scale depends on how far the query non-member sits from the training distribution.
Third, privacy and utility are read together: a low attack score certifies forgetting only when accuracy or mAP is preserved, whereas the same score reached while utility collapses, as for NegGrad+ and GA, is over-forgetting rather than removal.
Utility is always read against the reference named in each caption.
Fourth, every calibrated attack in this appendix is scored against a reference family that contains no checkpoint the core was distilled from and is not the row being evaluated, so no model is asked whether it is distinguishable from a population it was trained to imitate.
Fifth, on RegDB the retrained reference uses AdamW rather than the SGD schedule the ViT uses on the other two datasets, which does not train at all on the $824$ retained images; the baselines there are searched over their own rates for the same reason, so that no row reports a recipe failure as a method result.
Sixth, wherever the parasitic branch is read as a target, it is ranked against the other four seeds' cores and not against all five.
The target is one of the family's own members read a second way, so including its own core would compare it with itself; leaving it out puts the floor of the scale at $\tfrac{1}{10}$ rather than $\tfrac{1}{12}$, which is why the payload readings bottom out at $0.100$.

\FloatBarrier
\subsection{Training configuration}
\label{supp::impl}
The body gives the recipe; the remaining settings are here so the runs can be rebuilt without the code.
The backbone is ViT-B/16 pre-trained on ImageNet, following TransReID including random erasing~\cite{zhong2020random}.
Training uses SGD at $8\times10^{-3}$ under a cosine schedule with a parasitic learning rate multiplier of $5$ and batch $32$ on one NVIDIA RTX 5090, for $120$ epochs on Market-1501 and RegDB and $40$ on VeRi-776, in each case ending near the point where the core stops improving.
Step-1 distillation uses a feature-space objective against retain-only teachers, four of them on Market-1501, three on VeRi-776 and five on RegDB, weighted $0.15$ on Market-1501 and RegDB and $0.45$ on VeRi-776.
Step~1 takes one core update per iteration on Market-1501 and RegDB and one every fourth on VeRi-776, the period chosen so the core makes a comparable number of passes over its retain set each epoch: $372$ steps over $47$ retain batches is $7.9$ on Market-1501, $1154$ over $36$ at a period of four is $8.0$ on VeRi-776, and $129$ over $26$ is $5.0$ on RegDB.
A standalone retrain makes one pass per epoch.
Those extra passes are not extra fitting on Market-1501, where the reference already reaches $99.8$ mAP on its own retain set, and they are on VeRi-776, where it reaches $52.7$.
Both steps clip the gradient norm at $5$ and clamp every parasitic parameter to $[-5,5]$.
The Core-only control is the deployed configuration with \texttt{MOME.CORE\_ONLY} set, which disables Step~2 and changes nothing else: the two resolved configurations agree on every key that both record.
Across the five seeds the Market-1501 core holds $74.2$ mAP with a spread of $0.17$ and the retrained reference $70.5$ with $0.82$, so the $3.7$ between them is more than four times the wider of the two.
Running the consolidation stage on that Core-only core, with the same original and the same schedule the deployed payload uses, reaches $88.73$ mAP against the deployed payload's $88.7$ and leaves the Core it was built from untouched at $74.21$.
The two payloads also read alike under attack, $0.9977$, $0.9406$ and $0.8474$ against $0.9978$, $0.9419$ and $0.8511$, so the branch that ships does not depend on Step~2 having run.
The substitution is not free everywhere: on RegDB it reaches $65.7$ against the deployed payload's $66.6$, again leaving its Core untouched at $55.2$.
The equality is a Market result and what holds on both is that nearly all of the payload survives being built from a core that never co-trained.
Consolidation adds $30$ epochs at $4\times10^{-3}$ with label smoothing off, matching the objective that produced the original model.
BatchNorm running statistics are left free during Step 2 in every run reported here.
On the LayerNorm backbone that is safe and measured; on the BatchNorm ones it leaves the path open.
On RegDB the retrained reference and the ensemble that teaches the core use AdamW~\cite{loshchilov2019decoupled} at $3\times10^{-4}$, since the SGD schedule used elsewhere does not train at all on the $824$ retained images.
The core itself trains under the SGD schedule at $8\times10^{-3}$ like the other two benchmarks.
The baselines there run their own rates under the same rule: $5\times10^{-4}$ to $8\times10^{-4}$ for NegGrad+, SalUn, Bad Teacher and SCRUB, AdamW at $3\times10^{-4}$ for Fine-tune, which retained more utility there, $54.8$ mAP from visible to thermal against $5.2$ under the shared schedule, and the shared rate for GA, which survives it.
SSD takes no gradient step, and LoTUS and MUNBa carry their authors' rates.

Every baseline starts from the same standard checkpoint, sees the same retain set and the same deletion request, and runs its published objective inside the TransReID data pipeline, with its own hyperparameters at the values its authors report.
The weights say so rather than the scripts: every baseline checkpoint sits closer to that standard checkpoint than to the ImageNet initialization, with one exception, the dampening method on RegDB, which sits $0.174$ from the one against $0.171$ from the other.
That method moves furthest of the baselines on Market-1501 and still leaves $145$ of the $211$ core tensors bitwise identical to the checkpoint.
Two rows are exceptions.
LOFT is a projection of the embedding and leaves the weights untouched, and the LoRA row is ours rather than a port: low-rank adapters on the MLP projections of every block, backbone frozen, fitted to $\mathcal{D}_r$, following the low-rank adapter line of~\cite{cha2025loku}, which targets language models and has no re-identification counterpart to port.
One row is worth reading across splits rather than down a column: on RegDB, LoRA reproduces the original model to two decimals on the deployed split and moves well away from it on the other four, from $42.5$ to $51.1$ against the original's $64.7$.
The deployed run is not a no-op either.
It moves $66$ backbone tensors, the projections it adapts, by $2.3\times10^{-3}$ in relative norm, which is below what any column here resolves, where the other four move three times as far.
How far the adapter travels depends on which identities it is asked to forget, which is why the spread on that row is the widest in the table.
Methods that take their learning rate from the shared schedule are given a search over three rates where that schedule destroys the model, $10^{-3}$ down to $10^{-4}$ for SCRUB and the shared $8\times10^{-3}$ down to $5\times10^{-4}$ for NegGrad+, since the from-scratch rate is not meant for a converged ReID network.
On Market-1501 that is SCRUB and NegGrad+, and they are the two the sweep covers; Fine-tune, GA, SalUn and Bad Teacher keep the shared rate, which they survive.
LoTUS at $10^{-5}$, MUNBa and LOFT carry their own rates from their papers and run at those, and SSD dampens weights without a gradient step.
The main table reports the configuration that keeps the most utility, on the principle that a method which has destroyed the model cannot be said to have unlearned, and \cref{tab::lrsweep} gives the sweep.
Both swept methods trace the same curve.
NegGrad+ reaches the retrained reference on the attacks only at the rate that costs most of the retrieval, $28.2$ mAP against the reference's $70.5$, and SCRUB does not reach it at any rate we tried; both keep retrieval only where the attacks rise to the level of a model that never unlearned.
Neither end is the frontier, and the two ends belong to one method rather than to two, which is the same split the main comparison reports across methods.

\begin{table}[htbp]
\centering
\caption{\textbf{Learning rate traces the trade-off within a method} on Market-1501, seed $42$, identity-level deletion.
Each row is the same algorithm at a different rate.
The retrained reference and Core are repeated for scale, on the deployed split like every other row here, which is why Core reads $74.3$ and $80.1$ and the reference $70.5$ and $77.3$, against the five-split $75.3$, $80.5$, $71.4$ and $77.3$ of \cref{table::merged_market_veri}.
RMIA is the calibrated attack on $\mathcal{U}$; the reference reads $0.51$.}
\label{tab::lrsweep}
\setlength{\tabcolsep}{1.2mm}
\begin{tabular}{llccc}
    \toprule
    Method & Rate & test mAP & $\mathcal{U}$ mAP & RMIA \\
    \midrule
    SCRUB & $10^{-3}$ & 31.3 & 45.6 & 0.560 \\
          & $5\times10^{-4}$ & 43.9 & 65.2 & 0.709 \\
          & $10^{-4}$ & 85.4 & 99.7 & 0.987 \\
    \midrule
    NegGrad+ & $8\times10^{-3}$ & 28.2 & 31.7 & 0.494 \\
             & $10^{-3}$ & 86.2 & 99.7 & 0.990 \\
             & $5\times10^{-4}$ & 86.4 & 99.7 & 0.990 \\
    \midrule
    \textcolor{blue}{Retrain} & & \textcolor{blue}{70.5} & \textcolor{blue}{77.3} & \textcolor{blue}{0.511} \\
    \textbf{\textcolor{red}{MoME-Core}} & & \textbf{\textcolor{red}{74.3}} & \textbf{\textcolor{red}{80.1}} & \textbf{\textcolor{red}{0.511}} \\
    \bottomrule
\end{tabular}
\end{table}

\FloatBarrier
\subsection{Definitions of the reported privacy columns}
\label{supp::metricdefs}
Let $\boldsymbol{\theta}$ be the model under audit, $f(x;\boldsymbol{\theta})\in\mathbb{R}^{d}$ the embedding it retrieves with, and $p(x;\boldsymbol{\theta})\in\Delta^{C-1}$ the softmax of its identity head over the $C$ training identities.
$\mathcal{U}$ is the forget set and $\mathcal{Q}$ the non-member pool, the test query split.
The references $\boldsymbol{\theta}^{(1)},\dots,\boldsymbol{\theta}^{(K)}$ are retrains of the same retain set under different optimization seeds, none of which sees $\mathcal{U}$.

Two per-sample losses appear below.
The true-label loss $\ell(x,y;\boldsymbol{\theta})=-\log p_y(x;\boldsymbol{\theta})$, which is the statistic the loss-threshold attack scores~\cite{yeom2018privacy}, is defined only where the identity has a column in the head, which excludes $\mathcal{Q}$.
The loss-based columns therefore use the label-free
\begin{equation}
    \tilde\ell(x;\boldsymbol{\theta}) \;=\; -\log \max_{c} \, p_c(x;\boldsymbol{\theta}),
    \label{eq::selfloss}
\end{equation}
the cross-entropy of $x$ against the model's own prediction, which is the same quantity on both pools.
Every attack column is the area under the ROC curve of a score $s$ that is meant to be larger on members,
\begin{equation}
\begin{split}
    \mathrm{AUC}(s)=\frac{1}{|\mathcal{U}||\mathcal{Q}|}\sum_{x\in\mathcal{U}}\sum_{x'\in\mathcal{Q}}\Big(
    &\mathbf{1}\big[s(x)>s(x')\big] \\[-2pt]
    &+\tfrac12\mathbf{1}\big[s(x)=s(x')\big]\Big).
\end{split}
\label{eq::auc}
\end{equation}

\noindent\textbf{$\mathcal{U}$ mAP.} Retrieval is run inside $\mathcal{U}$, each image once as a query and once as a gallery entry.
For a query $x$ the gallery is $\mathcal{G}(x)=\{x'\in\mathcal{U}: x'\neq x,\ \neg(\,\mathrm{id}(x')=\mathrm{id}(x)\wedge \mathrm{cam}(x')=\mathrm{cam}(x)\,)\}$, ranked by cosine distance $1-\hat f(x)^{\!\top}\hat f(x')$ on $\hat f=f/\|f\|_2$, equivalently the Euclidean distance on $\hat f$, which the implementation uses.
With $r_x(k)=1$ when the $k$-th ranked entry shares the identity of $x$ and $R_x=\sum_k r_x(k)$,
\begin{equation}
    \mathrm{mAP}_{\mathcal{U}}=\frac{1}{|\mathcal{U}|}\sum_{x\in\mathcal{U}}\frac{1}{R_x}\sum_{k=1}^{|\mathcal{G}(x)|} r_x(k)\,\frac{1}{k}\sum_{r\le k} r_x(r).
    \label{eq::umap}
\end{equation}

\noindent\textbf{MIA$_\text{ent}$.} The uncalibrated entropy attack, one of the families surveyed in~\cite{hu2022membership}, scores confidence directly, with no references,
\begin{equation}
    s_\text{ent}(x)=-\mathcal{H}\big(p(x;\boldsymbol{\theta})\big)=\sum_{c=1}^{C} p_c(x;\boldsymbol{\theta})\log p_c(x;\boldsymbol{\theta}).
    \label{eq::miaent}
\end{equation}

\noindent\textbf{RMIA.} The target's loss is compared against the loss the reference population assigns to the same sample,
\begin{equation}
    s_\text{RMIA}(x)=\frac{1}{K}\sum_{k=1}^{K}\tilde\ell\big(x;\boldsymbol{\theta}^{(k)}\big)-\tilde\ell(x;\boldsymbol{\theta}).
    \label{eq::rmia}
\end{equation}

\noindent\textbf{U-LiRA.} The offline variant standardizes the same difference by the spread of the reference population.
With $\mu(x)$ and $\sigma(x)$ the mean and standard deviation of $\{\tilde\ell(x;\boldsymbol{\theta}^{(k)})\}_{k=1}^{K}$ and $\epsilon=10^{-6}$,
\begin{equation}
    s_\text{U-LiRA}(x)=-\frac{\tilde\ell(x;\boldsymbol{\theta})-\mu(x)}{\sigma(x)+\epsilon}.
    \label{eq::ulira}
\end{equation}

\noindent\textbf{Choice of the label-free loss.} The choice is forced rather than preferred, and the size of the artifact says why.
Scoring $\mathcal{Q}$ against a fixed dummy column instead, so that the true-label loss is defined everywhere, moves the calibrated columns further than the quantity they are meant to resolve and moves them differently on each benchmark.
On Market-1501 it puts Core at $0.467$ RMIA and the retrained reference at $0.446$, both below the null.
On RegDB it puts Core at $0.976$ against the reference's $0.471$.
A statistic that reads $0.47$ on a model retrained without $\mathcal{U}$ is not measuring membership, it is measuring where the deleted identities' head columns used to be, and the substitution moves the reference as far as it moves Core.
Under \cref{eq::selfloss} the two sit within $0.016$ of each other on all three benchmarks.

\noindent\textbf{Adaptation and power of the two calibrated attacks.} \cref{eq::rmia} and \cref{eq::ulira} are adapted forms, and we name them after the attacks they follow rather than claim to reimplement them.
RMIA in its original form is a pairwise likelihood ratio between the target and the reference models, calibrated over a population of reference samples; the form here keeps the reference calibration but reduces the comparison to the difference between the target loss and the reference mean.
The offline LiRA fits the reference population and reports a tail probability; the $z$ score of \cref{eq::ulira} is monotone in that probability under the same Gaussian, so it induces the same ordering, but the per-sample statistic is the label-free loss of \cref{eq::selfloss} rather than a logit-scaled true-label confidence, because the non-member pool has no column in the head.
The adaptations retain their power on this task, which is the property that matters here: against the same references, a model that did train on $\mathcal{U}$ is separated from a retrain at up to $0.38$ true positives for $1\%$ false positives against the reference's $0.012$, and the consolidated payload is detected on all three benchmarks.
An attack weakened into uninformativeness would not do that.

\noindent\textbf{The payload against the original model.} Consolidation initializes the branch at the difference toward the original and refines it, which invites the reading that Full is the original recovered by arithmetic and that the pair is therefore two copies of models that already exist.
It is not.
Measured over the $206$ tensors the two checkpoints share, the payload sits $0.071$ from the original in relative parameter norm where Core sits $0.099$, and $66\%$ of the payload's displacement from Core is orthogonal to the direction of the original, so refinement moves it off that line rather than along it.
Functionally it reaches full-data accuracy by a different route: $88.7$ mAP on Market-1501, above the $88.0$ of the Original row, from a point that is not on the line toward the model it was initialized from.
The stage leaves Core alone.
Of the $193$ core-side tensors the two files share, $188$ are bitwise identical, including all ten Batch Normalization running estimates.
The five that differ are the BNNeck batch counters, which advance because the stage runs forward passes and which evaluation does not read.
What a pair of separately trained files does not have, whatever it stores, is any statement tying the deletion to a property of the parameters that ship.
Core carries one and a retain-only student trained by some other procedure does not.

\noindent\textbf{Resolution of an attack AUC on these benchmarks.} Two numbers being close is not an equivalence test, and an interval computed over images is not the right interval, since images of one identity share a subject, a camera set and a difficulty.
Every interval here resamples identities.
Market-1501 offers $1126$ identity clusters across the two pools, VeRi-776 $488$ and RegDB $309$, and that difference decides what each benchmark can say.

\begin{table}[htbp]
\centering
\caption{\textbf{Attack AUC with the identity as the resampling unit.} RMIA on the label-free statistic against the disjoint reference family, $95\%$ interval from a bootstrap over identities at $2000$ resamples, and the difference from the retrained reference with its $90\%$ interval, which is the interval form of the two one-sided tests.
A difference whose interval contains zero is not resolved by the data; it is not thereby shown to be zero.}
\label{tab::clustered}
\setlength{\tabcolsep}{0.5mm}
\small
\begin{tabular}{l l c c c}
    \toprule
    & Row & AUC & $95\%$ int. & difference, $90\%$ int. \\
    \midrule
    \multirow{8}{*}{\rotatebox{90}{\textbf{Market}}}
    & \textcolor{blue}{Retrain} & \textcolor{blue}{0.511} & \textcolor{blue}{$[.491,.529]$} & \textcolor{blue}{--} \\
    & Core-only & 0.511 & $[.490,.534]$ & $+.001\ [-.021,+.023]$ \\
    & \textcolor{red}{Core s42} & \textcolor{red}{0.511} & \textcolor{red}{$[.488,.533]$} & \textcolor{red}{$+.000\ [-.022,+.023]$} \\
    & \textcolor{red}{Core s123} & \textcolor{red}{0.505} & \textcolor{red}{$[.484,.527]$} & \textcolor{red}{$-.005\ [-.027,+.017]$} \\
    & \textcolor{red}{Core s456} & \textcolor{red}{0.513} & \textcolor{red}{$[.492,.535]$} & \textcolor{red}{$+.003\ [-.019,+.026]$} \\
    & \textcolor{red}{Core s789} & \textcolor{red}{0.507} & \textcolor{red}{$[.485,.529]$} & \textcolor{red}{$-.003\ [-.025,+.020]$} \\
    & \textcolor{red}{Core s1024} & \textcolor{red}{0.512} & \textcolor{red}{$[.490,.534]$} & \textcolor{red}{$+.001\ [-.022,+.025]$} \\
    & \textcolor{gray}{Full} & \textcolor{gray}{0.942} & \textcolor{gray}{$[.930,.952]$} & \textcolor{gray}{$+.431\ [+.413,+.450]$} \\
    \midrule
    \multirow{8}{*}{\rotatebox{90}{\textbf{RegDB}}}
    & \textcolor{blue}{Retrain} & \textcolor{blue}{0.511} & \textcolor{blue}{$[.467,.554]$} & \textcolor{blue}{--} \\
    & Core-only & 0.516 & $[.470,.561]$ & $+.006\ [-.044,+.054]$ \\
    & \textcolor{red}{Core s42} & \textcolor{red}{0.526} & \textcolor{red}{$[.481,.571]$} & \textcolor{red}{$+.016\ [-.031,+.063]$} \\
    & \textcolor{red}{Core s123} & \textcolor{red}{0.569} & \textcolor{red}{$[.524,.614]$} & \textcolor{red}{$+.059\ [+.011,+.106]$} \\
    & \textcolor{red}{Core s456} & \textcolor{red}{0.531} & \textcolor{red}{$[.487,.576]$} & \textcolor{red}{$+.021\ [-.024,+.065]$} \\
    & \textcolor{red}{Core s789} & \textcolor{red}{0.557} & \textcolor{red}{$[.511,.601]$} & \textcolor{red}{$+.046\ [-.000,+.093]$} \\
    & \textcolor{red}{Core s1024} & \textcolor{red}{0.554} & \textcolor{red}{$[.506,.601]$} & \textcolor{red}{$+.043\ [-.006,+.094]$} \\
    & \textcolor{gray}{Full} & \textcolor{gray}{0.842} & \textcolor{gray}{$[.806,.874]$} & \textcolor{gray}{$+.331\ [+.289,+.374]$} \\
    \bottomrule
\end{tabular}
\end{table}

\cref{tab::clustered} says three things.
The interval half-width near the null averages $0.022$ on Market-1501 and $0.024$ on VeRi-776, whose smaller identity count is offset by a member pool of $18295$, and $0.045$ on RegDB, so a difference below roughly $0.04$, $0.05$ and $0.09$ is not resolved by these pools at all, whatever the point estimates look like.
On Market every Core row and the Core-only control differ from the retrained reference by at most $0.005$ with an interval that contains zero, while the consolidated payload differs by $0.431$ with an interval far from it, so the instrument separates what it should and does not separate Core from a retrain.
On RegDB one of the five seeds, $123$, excludes zero at the $90\%$ level before any allowance for having tested five; a second, $789$, has an interval whose lower end sits on zero at $-0.0004$, and the other three contain it comfortably.

That last reading is against one retrain, and one retrain is itself a draw.
Ranking each member of the disjoint family against the other four gives what retrains span among themselves: on RegDB $0.5016{\pm}.0323$ on RMIA, from $0.4684$ to $0.5499$, against Core's five seeds at $0.5475{\pm}.0180$, from $0.5262$ to $0.5689$.
Core's mean sits $1.43$ of those standard deviations above the family's and the two ranges overlap over most of their length, with one family member at $0.5499$ reading higher than two of the five Core seeds.
The same comparison on Market-1501 puts Core $0.73$ standard deviations above the family on RMIA and $1.66$ below it on the true positive rate.
So the separation a single reference showed is that reference being one draw from a spread of $0.032$, not Core being distinguishable from retraining, and the honest statement is that Core sits inside the band on both benchmarks and above its centre on each, by $0.73$ standard deviations on Market-1501 and $1.43$ on RegDB.

\noindent\textbf{Targets and references under different preprocessing.} Two rows on VeRi-776 need care, the original model and the LOFT projection applied to it.
The original was trained with a pixel normalization of $0.5$ in every channel while the retrained references use the ImageNet statistics, and an attack that pushes every model through one dataloader cannot serve both: under the target's pipeline the references are out of distribution, and under the references' pipeline the target retrieves at $36.5$ mAP instead of $82.4$.
The attack compares each model's own loss on a sample, and an adversary holding these models would run each one as it was trained, so the reference pass is given its own preprocessing.
The two passes keep the sample order fixed, so they align by index.
It decides five cells across those two rows.
Read through one pipeline the original model appears to leak at $0.60$ on the entropy attack, and read correctly it leaks at $1.00$, which is where a model that trained on every deleted identity belongs.

The control for the per-model pass is a second original trained from scratch on the same recipe with only the normalization changed, so that target and references share one input space and no per-model pass is needed.
It retrieves at $97.0$ Rank-1 and $79.6$ mAP against the released original's $97.1$ and $82.3$, and against the same four references it reads $0.99$, $0.92$ and $0.83$ on the three attacks, with $0.24$ true positives at $1\%$ false positives.
The row the table prints reads $1.00$, $0.91$ and $0.82$, with $0.25$.
The two differ by at most $0.018$ on the three attacks and by $0.012$ on the true positive rate, while the single-pipeline reading of $0.60$, $0.59$ and $0.58$ agrees with neither.
The per-model pass recovers what a natively matched model shows.

\noindent\textbf{The reference population against the distillation teachers.} Step~1 distills from an ensemble of retrains on $\mathcal{D}_r$, and the calibrated attacks score a target against a population of retrains on the same $\mathcal{D}_r$.
In the runs behind \cref{table::merged_market_veri} those two sets were the same four checkpoints, which makes the comparison circular: a model trained to imitate a population is then asked whether it is distinguishable from it.
To remove the overlap we trained four more retrains on the identical deletion split, none of which appears in any teacher list and none of which is the evaluated retrain row, and rescored every Market-1501 row against them.

\begin{table}[htbp]
\centering
\caption{\textbf{Rescored against references that share no checkpoint with the teachers.} Market-1501, identity-level deletion, members $\mathcal{U}$ against the query split.
The first pair of columns scores against the four retrains that Step~1 also distilled from; the second scores against four fresh retrains on the same split.
Entropy uses no references and is given once.}
\label{tab::disjointrefs}
\setlength{\tabcolsep}{0.9mm}
\begin{tabular}{l c cc cc}
    \toprule
    & & \multicolumn{2}{c}{teacher refs} & \multicolumn{2}{c}{disjoint refs} \\
    \cmidrule(lr){3-4}\cmidrule(lr){5-6}
    Row & ent & RMIA & U-LiRA & RMIA & U-LiRA \\
    \midrule
    Original & 0.999 & 0.991 & 0.979 & 0.991 & 0.979 \\
    Fine-tune & 0.982 & 0.984 & 0.975 & 0.984 & 0.974 \\
    \textcolor{blue}{Retrain} & \textcolor{blue}{0.493} & \textcolor{blue}{0.511} & \textcolor{blue}{0.514} & \textcolor{blue}{0.511} & \textcolor{blue}{0.511} \\
    \midrule
    Core-only & 0.491 & -- & -- & 0.511 & 0.515 \\
    \textcolor{red}{Core s42} & \textcolor{red}{0.492} & \textcolor{red}{0.510} & \textcolor{red}{0.503} & \textcolor{red}{0.511} & \textcolor{red}{0.513} \\
    \textcolor{red}{Core s123} & \textcolor{red}{0.490} & \textcolor{red}{0.504} & \textcolor{red}{0.498} & \textcolor{red}{0.505} & \textcolor{red}{0.510} \\
    \bottomrule
\end{tabular}
\end{table}

\cref{tab::disjointrefs} gives the result.
The objection predicts that removing the teachers from the reference population should push Core away from the null and expose a leak the overlap had hidden.
U-LiRA moves by $0.010$ and $0.012$ on the two Core rows, landing at $0.513$ and $0.510$ against the retrained reference's $0.511$; RMIA moves by a single unit in the last place on both.
Entropy reproduces to four decimals, which fixes the pipeline rather than the references.
The overlap was not carrying the result.

The rescoring covers all three benchmarks and every attack cell that carried a reading under the teacher references.
Of the $132$ such cells outside the VeRi-776 original, whose preprocessing is a separate question, thirty-eight change at the precision the tables print, twenty-nine by a single unit in the last place, and none by more than $0.03$.
The largest move is the retrained reference on RegDB, from $0.48$ to $0.51$, and it is a correction: the old comparison ranked that reference inside its own family, which pulls a member slightly below the null, while against an independent family it sits on the null, where a reference belongs.
Core on RegDB follows it from $0.50$ to $0.53$, so the gap between the two rows is what it was.
No conclusion in the paper turns on any of these moves.

The disjoint family also supplies its own band.
Training a fifth member on each benchmark and ranking each of the five against the other four gives the spread a genuine retrain has against its own kind, and Core's five seeds are read against it in \cref{tab::bandgap}.

\begin{table}[htbp]
\centering
\caption{\textbf{Core against the spread retrains have among themselves.} Each disjoint family has five members, each ranked against the other four, so the band is measured the way a target row is.
Every spread is a sample standard deviation over the five members of the family or the five seeds of Core.
The last column is Core's distance from the band's centre in units of the band's own standard deviation, and the one cell set in bold is the only one that falls outside $1.5$ of them in the direction that would mean a leak.}
\label{tab::bandgap}
\setlength{\tabcolsep}{1.0mm}
\small
\begin{tabular}{l l cc c}
    \toprule
    & Statistic & Retrain band & MoME-Core & distance \\
    \midrule
    \multirow{4}{*}{\rotatebox{90}{\textbf{Market}}}
    & MIA$_\text{ent}$ & $.4866{\pm}.0032$ & $.4914{\pm}.0015$ & $+1.5$ \\
    & RMIA & $.5000{\pm}.0132$ & $.5096{\pm}.0033$ & $+0.7$ \\
    & U-LiRA & $.4998{\pm}.0131$ & $.5135{\pm}.0021$ & $+1.0$ \\
    & TPR@$1\%$ & $.0112{\pm}.0014$ & $.0090{\pm}.0009$ & $-1.7$ \\
    \midrule
    \multirow{4}{*}{\rotatebox{90}{\textbf{VeRi}}}
    & MIA$_\text{ent}$ & $.5110{\pm}.0072$ & $.5195{\pm}.0043$ & $+1.2$ \\
    & RMIA & $.4994{\pm}.0141$ & $.5078{\pm}.0076$ & $+0.6$ \\
    & U-LiRA & $.4995{\pm}.0141$ & $.5058{\pm}.0090$ & $+0.4$ \\
    & TPR@$1\%$ & $.0096{\pm}.0019$ & $.0156{\pm}.0021$ & $\bm{+3.1}$ \\
    \midrule
    \multirow{4}{*}{\rotatebox{90}{\textbf{RegDB}}}
    & MIA$_\text{ent}$ & $.5187{\pm}.0204$ & $.5402{\pm}.0150$ & $+1.1$ \\
    & RMIA & $.5016{\pm}.0323$ & $.5475{\pm}.0180$ & $+1.4$ \\
    & U-LiRA & $.5002{\pm}.0321$ & $.5383{\pm}.0169$ & $+1.2$ \\
    & TPR@$1\%$ & $.0117{\pm}.0067$ & $.0177{\pm}.0059$ & $+0.9$ \\
    \bottomrule
\end{tabular}
\end{table}

Ten of the twelve cells place Core within $1.5$ standard deviations of the band.
Of the two that do not, the Market-1501 low false positive column sits $1.7$ below it, which is the over-forgetting direction and not a leak.
The one that matters is the same column on VeRi-776, where Core reads $.0156$ against the band's $.0096{\pm}.0019$, a distance of $3.1$.
In counts that is about $285$ of the $18295$ members recovered at $1\%$ false positives against the band's $176$, so roughly a hundred more, and we do not claim Core is indistinguishable from a retrain at that operating point on that benchmark.
Core's own spread is tighter than the family's everywhere, four times so on Market-1501 RMIA, $.0033$ against $.0132$, which is what a fixed retain set and a fixed teacher ensemble buy over training from scratch five times.
What this establishes is that the particular checkpoints did not matter, since the new references are themselves retrains on the same retain set.
It does not establish that distilling from retrained teachers is neutral in general, and the cost of that ensemble, four training runs, is stated where the method is described.

\noindent\textbf{Contribution of the retain-side recipe.} The lift over a plain retrain separates into the two parts the method claims for it.
Trained at seed $42$ with the arms differing only in the teacher list, Core reads $73.16$ mAP under the Step-1 pacing with no distillation, $74.36$ with one retain-only teacher and $74.89$ with four, against $70.5$ for a retrain on the same split.
The pacing carries $2.7$ of the $4.4$ point gap, the first teacher $1.2$ and the three after it $0.5$.
Read the last step with the drift between runs in mind: the four-teacher arm is the deployed configuration retrained alongside the others and it gives $74.89$ where the deployed run reads $74.34$, so half a point separates two runs of one configuration and this series resolves only the first two steps.

\noindent\textbf{The payload readout across backbones.} Full reads the same weights as Core with the parasitic branch on.
That readout did train on $\mathcal{U}$, so it is not exchangeable with the retain-only family, and on every backbone all five of its readings fall outside the range the reference family spans, which is the strongest separation an exact rank test on five against five can show, $p{=}0.008$.
Its direction follows how the branch fails and is not itself a privacy quantity.

\noindent\textbf{Omission of the representation-level probe.} A linear probe on hidden features is a standard membership attack and we ran it, then removed it from the tables because it does not measure membership under this protocol.
Let $h(x)\in\mathbb{R}^{d}$ be the class token at block $10$ of the frozen backbone and fit $s_\text{LP}(x)=\sigma(w^{\!\top}h(x)+b)$ to separate $\mathcal{U}$ from $\mathcal{Q}$.
Two facts sink it.

First, the folds have to be grouped by identity.
The two pools hold different identities and every identity contributes several images, so folds split by image place one identity on both sides and let the probe score by telling identities apart.
Grouping drops the Market-1501 column, over the unlearning rows and the reference, from a range of $0.78$ to $0.92$ to one of $0.63$ to $0.72$.

Second, what survives fails a positive control.
The consolidated payload encodes $\mathcal{U}$ by construction and every other attack detects it, entropy at $1.00$ and RMIA at $0.94$ on Market-1501.
The probe reads it at $0.607$ against the retrained reference's $0.635$, below the model that never saw the data.
On VeRi-776 the probe cannot separate the original model from the reference at all, $+0.003$, while fine-tuning, which leaks at entropy $0.93$, reads $0.032$ below the reference.
Only RegDB behaves, $+0.184$ for the payload and $+0.107$ for the original, and RegDB is the one benchmark whose two pools are the same size and drawn from one protocol.

The cause is the deletion unit rather than the probe.
Under identity-level deletion $\mathcal{U}$ and $\mathcal{Q}$ are disjoint identity sets of different sizes and difficulty, so separability between them is dominated by the pools themselves and membership is a small perturbation on top.
Grouping removes the shortcut of learning the identities; it cannot remove the difference between the pools.
An instrument that reads a knowledge-bearing model below the null is not reporting leakage, so we report the measurement and not the column.

\noindent\textbf{Reading the four columns.} They do not share an anchor, and only two of them have a meaningful absolute value.
$\mathcal{U}$ mAP is not read against zero: the reference never trained on $\mathcal{U}$ and still retrieves it far above chance, because identities are visually coherent and the embedding generalizes, so the quantity of interest is the distance to the reference's own $\mathcal{U}$ mAP.
MIA$_\text{ent}$ is uncalibrated and its scale depends on how far $\mathcal{Q}$ sits from the training distribution, so it too is read against the reference in the same column.
RMIA and U-LiRA subtract the reference population from the target by construction, so a genuine retrain sits at $\tfrac12$ and that value is meaningful: a score far below it is as distinguishable as one equally far above, since an adversary can invert its decision.
Finally, no column certifies removal alone.
An attack at the null reached while retrieval collapses is over-forgetting rather than unlearning, which is why utility and privacy are read together.

\FloatBarrier
\subsection{The threat model}
\label{supp::threat}
The guarantee is about one artifact under one release policy, and both have to be named before an attack result means anything.

\noindent\textbf{The released artifact.} Honoring a request means shipping $\bm{\theta}_c$ with the branch removed.
What a downstream party receives is the backbone and the retrieval feature read before the BNNeck, a $3840$-dimensional embedding on the ViT configuration.
The classification head is a training artifact.
It is not part of what is released, and the membership attacks in this paper read it anyway, so every attack number here scores a strictly larger object than the deployed one.

\noindent\textbf{Adversary capabilities.} The adversary holds $\bm{\theta}_c$ in full, white box, and may compute any function of it: embeddings for chosen images, logits through the head it was not given, gradients, and any statistic of the parameters.
It knows the deletion protocol, the architecture, the schedule, the sizes $|\mathcal{D}_r|$ and $|\mathcal{U}|$, and it may train its own reference models on data of its choosing.
This is what the calibrated attacks assume, and it is why they are run with reference families rather than against a fixed threshold.

\noindent\textbf{Artifacts withheld from the adversary.} The payload, the original model, and any intermediate checkpoint of the co-trained run.
A party holding the payload alongside Core recovers the deleted capability by design, which is the point of keeping the two separate.
Nothing here claims that a controller who retains the payload has deleted anything.
The claim is about $\bm{\theta}_c$: every input it depends on, the initialization, $\mathcal{D}_r$, the Step-1 randomness and the retain-only teachers, is free of $\mathcal{U}$, so releasing it discloses no more about $\mathcal{U}$ than releasing a model trained without it.

\noindent\textbf{Timing of the deletion request.} \cref{prop::indep} covers a set named before the core is trained.
A later request naming identities the core already learned is outside it, since the initialization the induction starts from would then depend on those identities.
Applying isolation to a trained model inherits nothing from the proposition, and \cref{tab:posthoc} shows it does not recover the behavior either: the core keeps the full-data accuracy and stays identifiable.

\noindent\textbf{Scope of the guarantee.} \cref{rem::scope} states the boundary formally.
The guarantee is relative to this training run.
Whatever a public pre-trained initialization already encodes about a requested identity is not something isolation can remove, and no claim here reaches it.
Nor does the guarantee say anything about $\mathcal{D}_r$, which Core legitimately learns.

\FloatBarrier
\subsection{The coupling channels, audited}
\label{supp::channels}
\noindent\textbf{A natural experiment for isolation.} Two rows of \cref{table::ablation_components} give the claim a test nobody has to design.
The $\bm{\gamma}$ only and $\bm{\beta}$ only variants change what the parasitic branch is made of while leaving the retain set, the seed, the schedule and the data order untouched, so isolation predicts that the core comes out the same.
Comparing the two checkpoints tensor by tensor, $201$ of the $211$ core tensors are bitwise identical.
The ten that differ are exactly the running mean and running variance of the five BNNeck modules, which are buffers written by the forward pass rather than parameters reached by a gradient.
Their difference is not small, up to $0.97$ in absolute value, and it changes nothing that ships: perturbing every running statistic in the network leaves the deployed retrieval feature unchanged to the last bit.
The two variants also read the same on the attack, $0.4873$ against $0.4873$, and differ only where they should, in Full mode at $78.1$ against $77.8$ mAP.

\noindent\textbf{Size of the admitted channel.} A4 does not close the random stream, it bounds what can cross it to a count of forget batches, and the size of that effect is measurable.
Training the same seed with Step~2 disabled gives a core that differs from the deployed one by $1.26\%$ of its norm over the $196$ core parameter tensors, while the BNNeck running statistics, which are buffers rather than parameters, differ by $73.8\%$.
For scale, the deployed core sits $12.9\%$ from a retrain on the same $\mathcal{D}_r$, and between $12.9$ and $13.2\%$ from each of that retrain's five seeds.
So everything Step~2 does to the core, which by construction is the random stream and nothing else, amounts to about a tenth of the distance between two models trained on the same data, and it does not move the reading: the two score $0.492$ and $0.491$ on the entropy attack and agree to $0.002$ on both calibrated ones.

\cref{ass::channels} names five conditions under which isolation is exact.
Those conditions are about paths from data to parameters, and a path can exist outside the gradient.
The list below names every path we could identify in this implementation and what closes it.
Two are closed by measurement rather than by construction, and both are reported.
The head Batch Normalization statistics are the subject of \cref{table::bn_freeze}.
The decision of whether to apply an update was originally taken once for both parameter groups, which is a content-carrying path, since whether a batch overflows depends on the batch; the decision is now taken per group, by testing each group's clipped gradient norm on its own.
It is not the scaler's decision and it does not depend on the scaler, which every run here constructs disabled since all of them train in bfloat16.
The loop counts each drop and reports the total once an epoch.
That counter ends at zero in every run that carries it, one of the Core-only runs among them.
The deployed runs predate the counter, so there the path is closed by the per-group decision and not by a log.
The classification head and the positional embeddings are core parameters, and what closes them is the freeze rather than a check, since Step~2 clears the gradient flag of the whole core group.
The one check inside the loop looks for the opposite leak, a parasitic gradient during Step~1, and it passes in all eighteen: twice per run where it is gated, and in the Core-only runs, where it is not, at every one of their $44640$, $11560$ and $15480$ iterations.
None of the loop's exceptional paths ran either.
The iteration that skips a non-finite Step-1 loss, the fallback to a core-only step on a non-finite Step-2 loss, and the mixed-precision backoff each log when taken, and none of those lines appears in the fifteen deployed runs or the three Core-only ones.
Every iteration of every run reported here took the path \cref{prop::indep} describes.

\begin{description}[leftmargin=0pt,itemsep=1pt,parsep=0pt,topsep=3pt,font=\normalfont\itshape]
\item[LayerNorm statistics.] Computed per sample, so no running buffers exist.
The built model carries $25$ buffers and every one of them sits on a BNNeck, across $31$ LayerNorm modules.
\item[Head BN statistics.] Buffers, so no gradient, and training normalizes by batch statistics, so no core gradient reads them.
Retrieval reads the feature before the BNNeck, and perturbing every running statistic leaves that feature bitwise unchanged, checked on the deployed checkpoint by shifting all ten running means and rescaling all ten variances, after which the $3840$-dimensional feature is equal bit for bit.
Freezing them in Step~2 keeps the buffers themselves free of $\mathcal{D}_f$; measured in \cref{table::bn_freeze}.
\item[Optimizer momentum.] Per-parameter buffers, so the core buffer sees Step-1 gradients only.
\item[Gradient clipping.] Clipping both groups together would leave Core's update depending on the size of the Step-2 gradient, since one factor rescales everything once the joint norm passes the threshold.
The two groups are clipped separately, at the same bound of $5$.
Two exceptional paths clip jointly and neither is a channel here: the Core-only control, where no Step~2 gradient exists to mix in, and the branch taken when the Step-2 loss is non-finite, which no run reported here entered.
\item[Mixed-precision scaler.] The reported runs use bfloat16, where no loss scaling is applied and the scaler is constructed disabled, so no scale is shared.
\item[Which update is applied.] Taken separately for the two parameter groups.
Whether a batch overflows depends on that batch, so a joint decision would carry content and not only a count.
A non-finite forget-batch gradient zeroes the parasitic group alone and the core update is applied as computed.
Across every checkpoint reported here the path was never taken.
\item[Dropout randomness.] Masks drawn independently of the input.
\item[Augmentation and sampling randomness.] Step 1 draws over $\mathcal{D}_r$ through its own loader; the two loaders share one pseudo-random stream, so the only quantity that can cross is how many forget batches were drawn, never their content.
\item[Activation cache.] Gradient checkpointing is off, so no Step-2 activation is replayed into a core backward pass.
\item[Classification head.] A core parameter; its columns are restricted to retained identities in Step~1 and it takes no gradient in Step~2.
The tensor is allocated at full width, so a requested identity's column exists and is never trained: on Market-1501 those $376$ rows carry $0.0235$ of norm against the retained rows' $0.5419$, and they are bitwise identical between the deployed run and the Core-only control, which share a seed and nothing else.
\item[Positional and side embeddings.] Core parameters, gradients only in Step~1.
\end{description}

\FloatBarrier
\subsection{Completeness of the normalization affine parametrization}
\label{sec::express}
\begin{proposition}[Completeness of the parasitic reparametrization]
\label{prop::express}
At any normalization site with output $\bm{\gamma}\odot\hat{\bm{x}}+\bm{\beta}$, every per-channel affine map $\bm{a}\odot\hat{\bm{x}}+\bm{b}$ is realized exactly by parasitic parameters $\bm{\gamma}_m=\bm{a}-\bm{\gamma}$ and $\bm{\beta}_m=\bm{b}-\bm{\beta}$.
Hence the Full family spans all per-channel affine reparametrizations of every augmented layer, a strictly larger hypothesis class than Core.
\end{proposition}
\noindent The completeness holds within the per-channel class only.
A normalization shadow rescales channels the frozen core already produces and cannot mix them, so any direction outside the core span stays unreachable no matter how the affines are set.
Low rank projection branches enlarge the reachable class and raise Full mode capacity during training, yet no branch trained from zero at any capacity we tried matches the original model on retrieval, the co-trained one reaching $82.0$ mAP on Market-1501 against the original's $88.0$.
The deployed payload is a different object, initialized at the exact parameter difference and then refined, and \cref{prop::express} delimits the normalization shadows alone rather than the branch as a whole, which also carries the low-rank pairs.

\FloatBarrier
\subsection{Low-rank structure of the parasitic gate}
\label{supp::svd}
\begin{figure*}[htp]
\centering
\includegraphics[width=\linewidth]{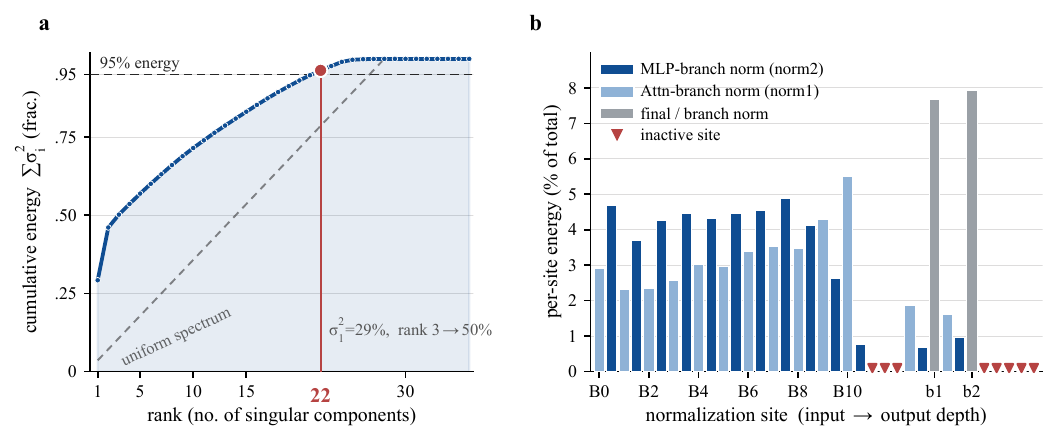}
\caption{\textbf{The parasitic gate $\bm{\gamma}_m$ is low-rank across normalization sites.} \textbf{Panel~a:} cumulative spectral energy of the stacked per-channel parasitic scales $\bm{\gamma}_m$ for the MoME model on Market-1501 at seed $42$ and epoch $120$, formed over $36$ normalization sites and $768$ channels, where $95\%$ of the energy lies in the top $22$ components, well above the uniform spectrum reference.
\textbf{Panel~b:} within the backbone the parasitic energy favors the normalization that precedes the MLP over the one that precedes attention, $54$ against $46\%$, in every block from $0$ to $8$, while blocks $9$ and $10$ invert it and the largest backbone site is the pre-attention norm of block $10$.
The two largest single sites are not in the backbone at all but are the final LayerNorms of the two JPM branches, ticked b1 and b2, at $7.7\%$ and $7.9\%$ against a backbone maximum of $5.5\%$.
Block $11$, the final norm and the five bottleneck sites are bypassed and carry exactly zero.}
\label{fig::svd}
\end{figure*}
\cref{prop::express} shows that the parasitic affine parameters are \emph{complete}: they can realize any per-channel gate at a normalization site.
Empirically the learned gate is also \emph{low-rank}.
Stacking the per-channel parasitic scales $\bm{\gamma}_m$ across all $36$ normalization sites of the Market-1501 MoME model into a $36\times768$ matrix and taking its SVD, we find that $95\%$ of the spectral energy lies in the top $22$ components, as \cref{fig::svd}a shows, far above a uniform spectrum; the result is stable across all five seeds, with $36$ sites, $r_{95}{=}22$ and the same eight silent sites in every one, and $\sigma_1^2$ between $29.1\%$ and $29.8\%$.
As \cref{fig::svd}b shows, within the backbone the energy concentrates in the normalization that precedes the MLP branch rather than the one that precedes attention, by $54$ to $46\%$ of the backbone energy, and every block from $0$ to $8$ puts more there while blocks $9$ and $10$ invert it, the single largest backbone site being the pre-attention norm of block $10$ at $5.5\%$; meanwhile block $11$, the final norm and the bottleneck sites are bypassed and carry exactly none.
The two largest single sites lie outside the backbone on every seed: the final LayerNorms of the two JPM branches hold $7.4$ to $7.9\%$ of the energy each, while no backbone site exceeds $5.5\%$, so the gate acts most strongly where the branch descriptors are normalized before retrieval.
The gap between the retain and full behaviors therefore occupies a compact, structured subspace, which is why the gating part of it is carried by the normalization affines.
Those affines are diagonal and cannot mix channels, so the deployed branch adds the zero-initialized low-rank pairs of \cref{sec::Method} on the linear maps.

This structure is learned, not imposed.
\cref{prop::express} guarantees only that the parasitic affines \emph{can} realize any per-channel gate; that the trained gate then collapses onto $22$ of the $36$ available singular directions, and onto the pre-MLP normalizations of the middle blocks, is a property of the retain-to-full gap itself.
It accounts for three observations at once.
The gate fits in the normalization affines because what it has to express is not spread across the backbone but confined to a compact subspace those affines span.
This is a statement about the gate the branch learns during co-training, not about the parameter difference to the original model: that difference is effectively full rank, which is why the deployed payload stores it densely rather than as a branch.
Widening the parasitic scope cannot move Core through the branch at all, since Step~2 never reaches it; the small movement that remains arrives through the shared random stream of \cref{ass::channels}, measured at $1.26\%$ of the core's norm.
What the scope changes is Full, and it changes it little, because the added directions fall almost entirely in the null space of that gap and carry no energy.
And the concentration in the middle blocks, with the JPM-bypassed final block empty, places that knowledge in the layers that route mid-level features rather than in the early or final stages.
The gate is thus complete in principle yet economical in practice, which is the property the switchable design relies on.

\subsection{Generality across backbones, pretraining and tasks}
\label{supp::generality}

\cref{tab:generality} gives the four settings behind the claim in the body, each against a reference retrained on its own $\mathcal{D}_r$: the deployed ViT, a ResNet-50 AGW backbone, CLIP-ReID, and a ResNet-18 trained from scratch for CIFAR-10 class deletion.
Core matches or beats its reference on utility on every row, by $3.7$ mAP on the ViT, $14.2$ on ResNet-50, $10.1$ on CLIP and a dead heat on CIFAR-10, where both score zero on the deleted classes at every seed.
The ResNet-50 margin is the largest, and neither side of it is seed noise: over five seeds the reference family spans $32.4$ to $34.5$ mAP and the core family $46.6$ to $48.7$.
What separates them is that the core distills from an ensemble of such references, which a single from-scratch retrain on a thin retain set has no way to do.

What the statistic resolves differs sharply across the four settings, and the reference column says by how much.
The retain-only family's own spread is $.085$ on the ViT and $.050$ on the ResNet-18, so a Core deviation of a few hundredths would be visible there, but it is $.241$ on ResNet-50 and $.232$ on CLIP, which is half the scale.
On those two backbones the Core reading carries no information about the forget set: a retrain against its own kind lands anywhere from $0.22$ to $0.82$, and Core sitting at $0.500$ inside that is not evidence.
The payload column is a different matter, since it leaves the band on every backbone.

The statistic separates the payload's two failure modes by direction.
The ViT payload restores $\mathcal{U}$ and lands at $0.100$, the memorizing direction.
On ResNet-50 and CLIP it does not carry $\mathcal{U}$ and deviates the other way, $0.880$ and $0.896$, and its retrieval on ResNet-50 is unstable rather than uniformly poor, $52.0 \pm 18.8$ mAP against a Core of $47.5 \pm 0.9$, collapsing on two seeds at $30.2$ and $32.8$ and recovering on three at $64.3$ to $67.7$.
On CIFAR-10 it lands at $0.152$, the memorizing direction, while its accuracy is chance, so what the branch carries is the forget set's loss profile and not a usable classifier.

\FloatBarrier
\subsection{Class deletion on image-classification benchmarks}
\label{supp::classbench}

\cref{tab:classbench} gives the class-deletion comparison behind the claim in the body: a tenth of the classes is removed, $\mathcal{U}$ is every image of those classes, and the table reports retained-class accuracy, accuracy on the deleted classes and entropy-MIA of $\mathcal{U}$ against the test split at three scales.
Every method is reimplemented against one ResNet-18, so all rows share a model, a split, a schedule and an evaluation, and every entry is a single run at seed $42$.

No single column separates the methods.
Fine-tune keeps retained accuracy and leaves the deleted classes recoverable, at $79.3\%$ on CIFAR-10 and $91.3\%$ on CIFAR-100.
GA and NegGrad+ reach the null only because the model is destroyed, at $11.1\%$ and $12.7\%$ retained accuracy against the reference's $94.3\%$.
SalUn falls below the null, $0.01$ against $0.26$, which is over-forgetting rather than privacy, and does not hold across scales: on TinyImageNet it leaves the deleted classes at $95.6\%$.
SCRUB is the one baseline that stays near the reference at all three scales, and no rule for combining the three columns is fixed in advance, so the comparison with it is column by column: on CIFAR-10 SCRUB is nearer on the attack, $0.24$ against the reference's $0.26$ where MoME reads $0.31$, while MoME is nearer on retained accuracy, $94.1\%$ against $93.8\%$ for a reference of $94.3\%$; on CIFAR-100 and TinyImageNet MoME matches the reference attack to $0.004$ and $0.002$, at $0.24$ and $0.22$, where SCRUB reads $0.28$ and $0.20$.
The remaining baselines do not order consistently as the class count grows.

\FloatBarrier
\subsection{Deletion requests with a different unit}
\label{supp::skewed}
\begin{table}[htbp]
\centering
\caption{\textbf{What a membership attack can and cannot verify} on Market-1501 at seed $42$.
Three deletion requests with different units, each trained from scratch, against the identity-level protocol above the rule.
The three skew rows run the base configuration, with no distillation ensemble and no low-rank pairs, while the row above the rule is the deployed recipe, so the comparison isolates the deletion unit only up to that difference.
Only identity skew removes whole identities, and it is the only row whose attack approaches the exchangeable value.
Camera skew and the imbalanced draw leave every identity in $\mathcal{D}_r$, so an attack asking whether an identity was trained on is answered yes however well those particular images were forgotten.
Each skew has its own forget set, so each row is scored against a reference family of four models retrained on that row's own $\mathcal{D}_r$, twelve retrainings in all.
All three attacks come from one evaluation per row against that family.}
\label{tab:skewed}
\setlength{\tabcolsep}{1.1pt}
\scriptsize
\begin{tabular}{lccccccc}
\toprule
Retain set skew & $|\mathcal{D}_r|$ & \multicolumn{2}{c}{Core} & Full & \multicolumn{3}{c}{Core MIA, AUC} \\
\cmidrule(lr){3-4}\cmidrule(lr){6-8}
 & & mAP & R-1 & mAP & MIA$_\mathrm{H}$ & RMIA & U-LiRA \\
\midrule
Uniform $\mathcal{D}_r$, identity-level & 1{,}491 & 74.3 & 88.2 & 82.0 & 0.492 & 0.511 & 0.513 \\
\midrule
Camera skew, delete cam-0            & 2{,}954 & 76.2 & 88.3 & 83.6 & 0.872 & 0.559 & 0.652 \\
Identity skew, delete 50 PIDs        & 12{,}025 & 87.5 & 94.5 & 86.0 & 0.541 & 0.517 & 0.518 \\
Imbalanced $k\!\in\![1,8]$     & 3{,}222 & 81.3 & 91.8 & 84.1 & 0.917 & 0.548 & 0.711 \\
\bottomrule
\end{tabular}
\end{table}

Everywhere else in this paper a request deletes whole identities: every image of a requested identity leaves, and $\mathcal{D}_r$ keeps $k$ images of each survivor.
A request need not arrive in that unit.
This section changes the unit and retrains MoME from scratch under each variant, all on Market-1501 at seed $42$.
\emph{Camera skew} deletes one fixed camera view per identity, so the request covers images rather than people and every identity is still present in $\mathcal{D}_r$.
\emph{Imbalanced} draws a per-identity retain count $k\!\sim\!U[1,8]$, so coverage varies from one image to eight and again no identity leaves.
\emph{Identity skew} deletes fifty identities outright and keeps every image of the survivors, the classical class-removal regime and the only variant here whose unit matches the main protocol; its retain set therefore grows to $12{,}025$ images.

The unit is what this table is about, because it decides what a membership attack is able to answer.

The retain set's size is a separate axis, and the main text reports the unthinned control on one run.
That run has an attack on it: with every image of the retained identities kept, so $6463$ against the deployed $1491$, the reference-free attack reads $0.4989$ on Core against the retrain's $0.4957$.
No reference family was trained for that protocol, so the calibrated attacks cannot be read there, and the pair is one run each at one seed.

\cref{tab:skewed} reports the outcome.
Core utility never falls below the uniform protocol: it reads $76.2$ mAP under the camera skew, $81.3$ under the imbalance, and $87.5$ under identity skew, where the far larger retain set simply gives the model more to learn from.
The attack column separates two things that are easy to confuse.
Identity skew deletes whole identities, and every attack on it sits near the exchangeable value, at $0.541$, $0.517$ and $0.518$.
That row also carries the only evidence here on the size of a request rather than its unit: it deletes fifty identities where the main protocol deletes $376$, and the two calibrated attacks read $0.517$ and $0.518$ against the main protocol's $0.51$ and $0.51$, so a request seven times smaller does not move them.
It runs the base configuration, without the distillation ensemble or the low-rank pairs, so it is not a like-for-like comparison with the deployed recipe.
Camera skew and the imbalanced draw leave every identity present in $\mathcal{D}_r$, and there the uncalibrated entropy attack reads $0.872$ and $0.917$, because an attack that asks whether an identity was trained on is answered yes however completely those particular images were forgotten.

Calibration is what separates the two questions.
Scored against a family retrained on the same skew's $\mathcal{D}_r$, which carries the same identities, that confound largely cancels: RMIA falls to $0.559$ and $0.548$ on the two identity-preserving skews.
What survives, up to $0.711$ on the imbalanced draw under U-LiRA, is the image-level signal, which is the part a deletion request over images can actually be held to.
MoME is thus robust to \emph{how} $\mathcal{D}_r$ is sampled, and an attack score means what the reference family lets it mean.

\FloatBarrier
\subsection{Resolution against request size}
\label{supp::reqsize}
The protocol everywhere else names half the identities.
A request under a right to erasure names far fewer, and retraining at every size is out of reach, so we answer the half of the question that the stored scores already settle.
Holding the deployed models fixed, we restrict the audit to a random subset of the requested identities, redraw the subset twenty times at each size, and record what the reading and its interval do.
This separates the resolution of the statistic from the behaviour of a model trained on a smaller request, which is a different question we do not answer, and it is the half a controller needs, since it says how many identities a request must name before a null reading carries information.

\cref{tab::reqsize} gives it.
Two readings.
The positive control survives all the way down: the consolidated payload is detected at every size, reading $0.926$ on five identities where the interval half-width is $0.066$, so a leak of that size is caught by a request naming five people.
The null reading does not survive nearly as far.
Core and the retrained reference stay together at every size, which is the result, but the interval around them widens from $0.022$ at the full request to $0.046$ at fifty identities and $0.153$ at five.
At five identities the audit cannot resolve a difference below about $0.3$, so a null there is not evidence of anything.
Reading the column against the $0.04$ the full request resolves, a request has to name on the order of a hundred identities before the calibrated attacks are as informative as the tables elsewhere in this paper.
The other two benchmarks give the same picture at their own scales.
On VeRi-776 the half-width runs from $0.025$ at the full $288$ identities to $0.115$ at five while the payload holds between $0.883$ and $0.897$; on RegDB, whose full request is only $103$ identities, it runs from $0.043$ to $0.155$ while the payload holds between $0.838$ and $0.863$.
On all three the reference and Core stay together at every size, so what the request size changes is the resolution and not the result.

The rank statistic degrades far more slowly, and the same subsetting shows it.
Its band, the spread of the retrained family's leave-one-out readings, widens only from $0.0845$ at the full request to $0.0863$ at fifty identities and $0.1260$ at five, a factor of $1.5$ where the attack interval widens by $7$.
Its positive control does not degrade at all: the consolidated payload read against the Core family sits at $0.100$ at every size down to five identities, which is the floor of the rank scale, so it stays $0.40$ from the null against a band of $0.126$, or $3.2$ of that band's standard deviations.
Part of that robustness is structural, since a control already at the floor of a bounded scale has nowhere to fall, but the band widening slowly is not: it is what makes the floor readable.
Where the calibrated attacks stop resolving anything at five identities, the rank test still separates a leaking readout from the null.
VeRi-776 repeats it more sharply: its band moves from $0.1013$ at the full $288$ identities to $0.1105$ at five, a factor of $1.1$, and the control sits at exactly $0.1000$ at every size.
That band is wider than Market-1501's in absolute terms, so the rank test resolves less there overall while losing almost nothing as the request shrinks.

\begin{table}[htbp]
\centering
\caption{\textbf{What the audit can resolve as the request shrinks.} RMIA on Market-1501 with the deployed models fixed, the audit restricted to a random subset of the requested identities, twenty redraws per size and $600$ resamples per draw.
The last row names all $376$, where the subset is the whole set and one draw is all there is.
The last column is the half-width of the identity-clustered $95\%$ interval on the Core row, which is the wider of the two rows near the null at every size.}
\label{tab::reqsize}
\setlength{\tabcolsep}{1.6mm}
\small
\begin{tabular}{l ccc c}
    \toprule
    Identities named & \textcolor{red}{Core} & \textcolor{blue}{Retrain} & \textcolor{gray}{Full} & half-width \\
    \midrule
    5   & \textcolor{red}{0.518} & \textcolor{blue}{0.534} & \textcolor{gray}{0.926} & 0.153 \\
    10  & \textcolor{red}{0.510} & \textcolor{blue}{0.522} & \textcolor{gray}{0.937} & 0.092 \\
    25  & \textcolor{red}{0.504} & \textcolor{blue}{0.518} & \textcolor{gray}{0.943} & 0.067 \\
    50  & \textcolor{red}{0.519} & \textcolor{blue}{0.507} & \textcolor{gray}{0.944} & 0.046 \\
    100 & \textcolor{red}{0.511} & \textcolor{blue}{0.509} & \textcolor{gray}{0.944} & 0.036 \\
    200 & \textcolor{red}{0.515} & \textcolor{blue}{0.515} & \textcolor{gray}{0.939} & 0.028 \\
    376 & \textcolor{red}{0.511} & \textcolor{blue}{0.511} & \textcolor{gray}{0.942} & 0.022 \\
    \bottomrule
\end{tabular}
\end{table}

\FloatBarrier
\subsection{Toward post-hoc unlearning}
\label{supp::posthoc}
\begin{table}[htbp]
\centering
\caption{\textbf{Applying isolation after training does not unlearn.} Post-hoc adaptation of the deployed checkpoint, Market-1501 at seed $42$, $30$ epochs, with the three reference rows above the rule.
The original here is the checkpoint these variants adapt, the one \cref{table::merged_market_veri} also reports.
Attacks are read in Core mode against the same four disjoint references as the main tables.
Every variant keeps the full-data accuracy and stays identifiable at close to the original's level whether or not \textit{gradient isolation} is applied, because isolation withholds new gradients and cannot remove what the weights already carry.
PH1 freezes the backbone, so its Core is the original model and it reproduces the original's three scores exactly.}
\label{tab:posthoc}
\setlength{\tabcolsep}{1.1pt}
\scriptsize
\begin{tabular}{lcccccc}
\toprule
Method & \multicolumn{2}{c}{Core} & Full & \multicolumn{3}{c}{Core MIA, AUC} \\
\cmidrule(lr){2-3}\cmidrule(lr){5-7}
 & mAP & R-1 & mAP & MIA$_\mathrm{H}$ & RMIA & U-LiRA \\
\midrule
Original, trained on $\mathcal{D}_\mathrm{full}$  & 88.0 & 94.8 & --   & 0.999 & 0.991 & 0.979 \\
Retrained reference: retrain on $\mathcal{D}_r$         & 70.5 & 86.9 & --   & 0.493 & 0.511 & 0.511 \\
MoME-Core, by design                    & 74.3 & 88.2 & 82.0 & 0.492 & 0.511 & 0.513 \\
\midrule
Naive post-hoc, no isolation                      & 88.5 & 95.2 & 88.8 & 0.997 & 0.967 & 0.909 \\
PH1: freeze backbone                              & 88.0 & 94.8 & 88.4 & 0.999 & 0.991 & 0.979 \\
PH2: isolation, low LR                           & 88.4 & 95.0 & 88.5 & 0.994 & 0.984 & 0.957 \\
PH3: isolation + LR restart                       & 87.8 & 94.7 & 88.4 & 0.967 & 0.968 & 0.927 \\
\bottomrule
\end{tabular}
\end{table}

MoME unlearns by design: Core is optimized from initialization and never sees $\mathcal{D}_f$.
A natural question is whether the same parasitic and core split can be \emph{retrofitted} onto a model already trained on the full data, the setting most unlearning methods target.
\cref{tab:posthoc} tests it over four schemes at $30$ epochs starting from the deployed checkpoint, and the answer is no.
All four keep the full-data accuracy, $87.8$ to $88.5$ mAP against the original's $88.0$ where the retrained reference reads $70.5$, and all four stay identifiable, $0.967$ to $0.999$ on the entropy attack and $0.967$ to $0.991$ on RMIA where the reference reads $0.493$ and $0.511$.
Gradient isolation does not change that.
It withholds new forget-set gradients and cannot remove what the weights already carry.

The four differ only in how far they are allowed to move the backbone, and none of them leaves its full-data solution in $30$ epochs.
\emph{PH1} makes this explicit by freezing the backbone: its Core is the original model, and it reads the original's three attack scores to four decimals, $0.9986$, $0.9907$ and $0.9786$.
\emph{PH2} and \emph{PH3} let the backbone move under isolation, at a low rate and at a restarted one, and both land within a point of where they started.
\emph{Naive}, which applies no isolation at all, is not distinguishable from the three that do.

This is what \cref{prop::indep} predicts rather than a failure of the schedule.
The induction that makes Core independent of $\mathcal{U}$ starts from an initialization that never saw it, and a core that begins as a function of the forget set stays one under any later restriction of its gradients.
A retrofit can reproduce the mechanism and not the guarantee, and here it does not reproduce the behavior either.
Both Full columns are the full-data model reached by a different route, and neither carries information the matching Core lacks, which is why the switch is not the interesting quantity in this table.

Repeating this across seeds and across the other two benchmarks is not done.
An earlier version of this study reported such a repetition, and those runs named the deployed checkpoint on the command line while leaving the initialization flag at its default, so they trained from ImageNet weights and measured a schedule rather than a retrofit.
They are withdrawn rather than reinterpreted.

\FloatBarrier
\subsection{A rejection region for the rank statistic}
\label{supp::ranktest}
The aggregate $\rho_{\mathcal U}$ over a leave-one-out family is a counting identity.
With $K$ shadows and no ties, the $K$ ranks a sample receives as each model takes its turn as target are a permutation of $\tfrac{2i-1}{2K}$, so their mean is exactly $\tfrac12$ whatever the family did.
Reading $\tfrac12$ back off that average is not evidence, and a test needs a unit, a rejection region and a calibration.

\noindent\textbf{The within-family reading.} Core is ranked among shadows of its own procedure.
Ranking it among the retrains it is compared against is the other reading, and it cannot separate them.
Ranking Core against the five retrained references on Market-1501 puts it at $0.083$ on every seed, which is the floor of the scale at six models, and ranking the consolidated payload against the same five puts it at $0.083$ as well.
The reading that is supposed to separate them cannot, because Core's loss on $\mathcal{U}$ is below all five references on every forgotten sample.

That is generalization and not memory, and the retain set says so.
On the Market-1501 images both procedures trained on, Core is above all five references rather than below them, at a mean loss of $0.313$ against $0.172$: the references fit their thin retain set harder than Core does, while Core is the better model on identities neither ever saw.
Throughout this subsection a loss is the deployed Core against the five-seed mean of the reference family.
A model carrying residual memory of $\mathcal{U}$ would be the more confident one on the data it kept, not the less confident one.
The cost of the within-family reading is real and we state it: a failure of isolation that affected the procedure itself would move every member together and the leave-one-out reading would not see it, which is why the power of the statistic is demonstrated with planted violations rather than claimed from the deployed number.

The unit is one model paired with one identity, since a violation is coherent inside an identity and nowhere else.
For model $t$ and identity $i$ contributing $n_i$ scored images let $m_{t,i}$ be the mean rank and
\begin{equation}
    z_{t,i}=\big(\tfrac12-m_{t,i}\big)\frac{\sqrt{n_i}}{\sigma_1},\qquad
    \sigma_1=\sqrt{\frac{K^2-1}{12K^2}},
    \label{eq::zcell}
\end{equation}
where $K$ is the number of models a sample is ranked among, so a single rank takes $K$ equally spaced values and $\sigma_1$ is their standard deviation.
A two-image identity at the floor and a fifty-image identity slightly below the middle are thereby placed on one scale, and under exact exchangeability $z$ has unit variance.
On Market-1501 with $K{=}5$ and the $376$ deleted identities this gives $1880$ cells per family.
This is a different spread from the one \cref{tab:generality} reports: there the unit is a whole model and the number is how far one seed's $\rho_{\mathcal U}$ falls from another's, here the unit is one identity inside one model.

\begin{table}[htbp]
\centering
\caption{\textbf{The analytic null is wrong, and the clean family says by how much.} Distribution of \cref{eq::zcell} over the $1880$ cells of each Market-1501 family.
The first two rows are leave-one-out within their own family, the third reads the consolidated payload against the Core references.
Exact exchangeability predicts unit variance, and that prediction assumes the ranks are independent across the images of one identity, which they are not.
A family of genuine retrains already reads $2.68$, so a rejection against the analytic null is not evidence of retained data, which is why every reading in this paper is calibrated against the retrained family rather than against $\tfrac12$.}
\label{tab::zcells}
\setlength{\tabcolsep}{1.5mm}
\begin{tabular}{l ccc c}
    \toprule
    Family & sd & p99 & max & above $3.8$ \\
    \midrule
    Retrain & 2.68 & 6.30 & 9.63 & 137 \\
    MoME-Core & 3.15 & 7.19 & 10.23 & 220 \\
    \midrule
    Full vs Core & 1.80 & 10.86 & 12.00 & 1520 \\
    \bottomrule
\end{tabular}
\end{table}

\cref{tab::zcells} reads as follows.
Both leave-one-out families have mean zero by the counting identity.
The retrained family, which by construction never saw the deleted identities, already carries a standard deviation of $2.68$ against the analytic $1$, and $137$ of its $1880$ cells sit beyond the nominal $5\%$ family-wise threshold of $3.8$.
What that measures is the correlation just described rather than a failure of \cref{prop::rank}: seeds land in different optima and an identity is easy or hard coherently across its images, so permuting the model axis independently per image destroys a dependence that is present under the null and makes the simulated null too narrow.
A permutation test against it rejects for the retrained family at $p\le0.005$, so that test is not usable and we do not use it.
MoME-Core reads $3.15$ and $220$ cells, the same order as the retrained family.

What separates a model that kept the data is of a different order.
Reading the consolidated payload against the Core references moves the whole distribution rather than its tail: the mean goes from $0$ to $5.59$ and $1520$ of $1880$ cells clear the threshold.
The instrument has power, and the gap between the two leave-one-out families is small next to the gap either has to the payload.
This is the same conclusion \cref{tab::rhopower} reaches with planted violations, reached here without planting anything.

\noindent\textbf{The families on all three benchmarks.} \cref{tab::rhothree} lists them.
The deployed Core reads $0.538$, $0.487$ and $0.494$, each inside the range its own retrained family spans, and the spread of that reference family is what sets the resolution on each benchmark.
Two things are worth reading off it.
The mean of every family is exactly $0.5000$, which is the counting identity and not a result.
And a family of from-scratch retrains spreads two to four times as wide as a family of MoME cores on every benchmark, which is the same instability the retrain shows on retrieval, so the band a deployed audit is read against is set by how unstable retraining is on a thin retain set.

\begin{table}[htbp]
\centering
\caption{\textbf{Forgetting-Rank families on all three benchmarks.} Each row gives the five leave-one-out readings of one family, deployed seed first, and their sample standard deviation.
Every family averages $0.5000$ by the counting identity, so the spread is the informative quantity. {\color{red}Red}: the deployed Core, whose own reading is in bold.}
\label{tab::rhothree}
\setlength{\tabcolsep}{1.2mm}
\small
\begin{tabular}{l ccccc c}
    \toprule
    Family & \multicolumn{5}{c}{leave-one-out $\rho_{\mathcal U}$} & sd \\
    \midrule
    \textcolor{blue}{Market retrain} & \textcolor{blue}{0.603} & \textcolor{blue}{0.525} & \textcolor{blue}{0.400} & \textcolor{blue}{0.427} & \textcolor{blue}{0.544} & \textcolor{blue}{.085} \\
    \textcolor{red}{Market Core} & \textcolor{red}{\textbf{0.538}} & \textcolor{red}{0.480} & \textcolor{red}{0.486} & \textcolor{red}{0.501} & \textcolor{red}{0.495} & \textcolor{red}{.023} \\
    \midrule
    \textcolor{blue}{VeRi retrain} & \textcolor{blue}{0.559} & \textcolor{blue}{0.354} & \textcolor{blue}{0.553} & \textcolor{blue}{0.437} & \textcolor{blue}{0.597} & \textcolor{blue}{.101} \\
    \textcolor{red}{VeRi Core} & \textcolor{red}{\textbf{0.487}} & \textcolor{red}{0.522} & \textcolor{red}{0.440} & \textcolor{red}{0.544} & \textcolor{red}{0.507} & \textcolor{red}{.040} \\
    \midrule
    \textcolor{blue}{RegDB retrain} & \textcolor{blue}{0.611} & \textcolor{blue}{0.392} & \textcolor{blue}{0.582} & \textcolor{blue}{0.487} & \textcolor{blue}{0.428} & \textcolor{blue}{.095} \\
    \textcolor{red}{RegDB Core} & \textcolor{red}{\textbf{0.494}} & \textcolor{red}{0.539} & \textcolor{red}{0.520} & \textcolor{red}{0.509} & \textcolor{red}{0.437} & \textcolor{red}{.039} \\
    \bottomrule
\end{tabular}
\end{table}

\FloatBarrier
\subsection{Membership inference diagnostics}
\label{supp::mia_diag}
The analyses below accompany the privacy results of \cref{sec::Experiments}: which member pool the signal comes from, where the retained accuracy comes from, the one buffer that could leak outside the gradient path, the low false positive region, the choice of a reference family over a single reference, and the Forgetting-Rank in full.

\emph{Which member pool the attack is asked about.} A single membership number mixes populations that behave differently, and only one of them is about removal.
\cref{table::mia_decomp} scores four pools against the same query non-member set.
The reference sets the pattern.
It reads $1.000$ on $\mathcal{D}_r$, the images it trained on, and $0.938$ on the images of those same identities that $\mathcal{D}_r$ did not keep, which it never saw: an identity the model learned makes its other images look like members, so a high score there is generalization rather than memorization.
On $\mathcal{U}$, whose identities the reference never met, it reads $0.493$, at chance.
Core reproduces all three, at $1.000$, $0.945$ and $0.493$.
The consequence is that the whole training set reads $0.722$ for the reference and $0.725$ for Core, far above chance for two models that forgot correctly, which is why a pool built as the complement of $\mathcal{D}_r$ cannot certify anything.
The failure cases frame the reading: Fine-tune stays at $0.982$ on $\mathcal{U}$ and gradient ascent above $0.99$ on every pool alike, identifiable everywhere rather than selectively forgetful.
Full, which by design encodes $\mathcal{D}_f$, reads $0.998$ on $\mathcal{U}$, which is where a knowledge-bearing readout belongs.

\noindent\textbf{Source of the retained accuracy.} The same three pools separate retrieval, and \cref{table::partition} reads them.
Core leads the retrained reference on $\mathcal{U}$ by $2.8$ on Market-1501 and $5.3$ on VeRi-776, and leads it by $3.9$ and $4.4$ on the retained identities it never trained on.
The two gaps move together, so the lead on $\mathcal{U}$ is the same generalization that shows up on data nobody asked to delete, not memorization of the deleted identities.
How much generalization there is to have is easy to understate from the mAP column alone: the reference's Rank-1 on $\mathcal{U}$ is $90.1$ on Market-1501 and $64.8$ on VeRi-776, so a model that never trained on these identities still puts another of their images first most of the time.
Over five seeds the Market gaps read $2.98$ and $4.08$ and the VeRi-776 gaps $8.36$ and $7.91$, so neither ordering is a seed artifact, and both benchmarks put the two gaps within $1.1$ of each other while the gaps themselves run from $3$ to $8$.
The left column is where the two differ most and it is not a privacy quantity: both fit $\mathcal{D}_r$ almost exactly on Market-1501, $100.0$ against $99.8$, while on VeRi-776 Core reads $99.9$ and the reference only $56.0$ at this seed, against $52.7$ over five.
A from-scratch retrain on $1152$ images does not converge on its own training set, which is the same instability its five seeds show as a $5.1$ spread in test mAP, and it is why the reference is weak on that benchmark rather than why Core is strong.

\begin{table}[htbp]
\centering
\caption{\textbf{Retrieval mAP on the three parts of the training set} at seed $42$, under identity-level deletion.
The retained, held out column holds images of retained identities that were never in $\mathcal{D}_r$, so high values there are generalization rather than leakage.
Full is the consolidated payload, the artifact reported in \cref{table::merged_market_veri}, not the parasitic branch as trained.
Every row here is one run on the deployed split, where \cref{table::merged_market_veri} averages five splits, which is why Core reads $80.1$ on $\mathcal{U}$ here against $80.5$ there while the Market reference reads $77.3$ in both.
The same gap is larger on VeRi-776, $35.5$ here against $32.2$ there for the reference, because that reference is the unstable one: its five seeds span $27.6$ to $39.5$ mAP.}
\label{table::partition}
\setlength{\tabcolsep}{1.3mm}
\begin{tabular}{l l ccc}
\toprule
Dataset & Method & $\mathcal{D}_r$ & retained, held out & $\mathcal{U}$ \\
\midrule
\multirow{3}{*}{\rotatebox{90}{\textbf{Market}}}
 & Retrain & 99.8 & 86.4 & 77.3 \\
 & \textbf{MoME-Core} & 100.0 & 90.3 & 80.1 \\
 & \textbf{MoME-Full} & 99.8 & 99.8 & 99.8 \\
\midrule
\multirow{3}{*}{\rotatebox{90}{\textbf{VeRi}}}
 & Retrain & 56.0 & 47.7 & 35.5 \\
 & \textbf{MoME-Core} & 99.9 & 52.1 & 40.8 \\
 & \textbf{MoME-Full} & 100.0 & 100.0 & 100.0 \\
\bottomrule
\end{tabular}
\end{table}

\emph{Batch Normalization side channel.} Gradient isolation controls what the gradients touch, but the classification heads carry Batch Normalization, whose running mean and variance are updated by the forward pass rather than by gradients.
A Step-2 pass over $\mathcal{D}_f$ could therefore in principle write forget-set statistics into a buffer that Core later reuses, a leak outside the gradient path.
\cref{table::bn_freeze} tests this by freezing those statistics during Step~2 and comparing against the default.
The membership attack on $\mathcal{U}$ moves by $0.006$, so the buffer carries no meaningful $\mathcal{D}_f$ signal and the LayerNorm-backbone configuration is safe without the freeze.
The channel is specific to Batch Normalization, which is why it is absent on the LayerNorm backbone yet must be closed on the BN-normalized CNNs of the generality study.
Those runs did not close it.
Their ResNet-50 carries one set of running buffers for fifty-five backbone sites, shadowed only at the BNNeck, so Step~2 wrote $\mathcal{D}_f$ statistics into buffers Core reads.
No privacy claim is made on that backbone for a different reason, that its band resolves nothing, and this is a second.

\begin{figure}[htbp]
\centering
\includegraphics[width=\linewidth]{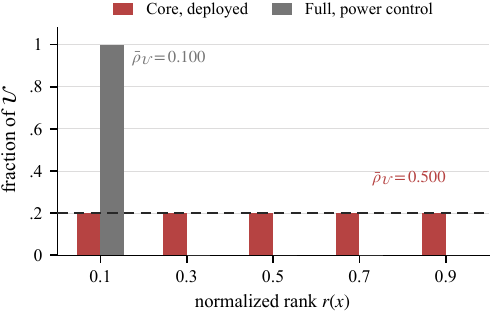}
\caption{\textbf{Calibration and power of the Forgetting-Rank on Market-1501} under identity-level deletion, over five shadows sharing one split, $|\mathcal U|{=}6473$ forgotten images and $32365$ scored ranks per series.
In red, the deployed Core ranked among the shadows that also forgot $x$: every level carries exactly $0.200$ of the mass, the dashed rule, at $\chi^2$ $p{=}1.0$ and $\bar\rho_{\mathcal U}{=}0.500$.
This is the null being exactly calibrated.
Because the ranks of one family are a permutation across its targets, that uniformity follows from a counting identity whatever the model did, so it certifies the reference rather than the forgetting.
In grey, the same network read in Full mode and scored against the other seeds' Core, a comparison across families that the identity does not constrain: it sits at the floor of the rank scale for $99.98\%$ of forgotten samples, $\bar\rho_{\mathcal U}{=}0.100$.
The reading does not depend on treating images of one identity as independent: with the identity as the unit, every one of the $1880$ identity readings the five models give falls below the null and $1877$ sit at the floor.
The test's power is located there.}
\label{fig::rankcert}
\end{figure}

\emph{The audit statistic in detail.} \cref{fig::rankcert} separates what the Forgetting-Rank calibrates from where its power lies.
The per-sample rank density of Core, in red, is flat, carrying exactly $0.200$ of the mass at each of the five levels at $\chi^2$ $p{=}1.0$, for a mean of $\rho_{\mathcal U}{=}0.500$.
That flatness is the null being exactly calibrated rather than a reading about Core: the ranks of one family are a permutation across its targets, so the pooled density is uniform by a counting identity whatever the model did, and the retrained reference reads the same $0.500$ for the same reason.
Power comes instead from comparing across families, which no identity constrains.
The identical network read in Full mode, in grey, whose parasitic branch was trained on $\mathcal{U}$, is scored against the other seeds' Core and sits at the floor of the rank scale for $99.98\%$ of forgotten samples, at $\rho_{\mathcal U}{=}0.100$.
Images of one identity are not independent, so the count that matters is taken over identities: of the $1880$ identity readings the five models give, every one falls below the null and $1877$ sit at the floor, with a spread across them of $0.007$.
This is why we report the statistic as a diagnostic with an explicit positive control rather than as the paper's privacy result, and why the test that carries a rejection region, \cref{supp::ranktest}, takes the identity rather than the family average as its unit.

\begin{table}[htbp]
\centering
\caption{\textbf{Membership inference AUC by member pool on Market-1501}, the entropy attack of \cref{eq::miaent}, under identity-level deletion at seed $42$.
The last three pools partition the training set and the first is their union: $\mathcal{D}_f$ is all $12936$ images, $\mathcal{U}$ the $6473$ images of the deleted identities, $\mathcal{D}_r$ the $1491$ images kept from the retained ones, and the last column the $4972$ images of retained identities that $\mathcal{D}_r$ did not keep.
The query split is the non-member set throughout.
Only the $\mathcal{U}$ column asks about removal.
The reference reads $1.000$ on the images it trained on and $0.938$ on retained identities it never trained on, so a pool that mixes them certifies nothing; it is at chance only on $\mathcal{U}$. {\color{red}Red}: Core; {\color{blue}blue}: retrained reference; \textcolor{gray}{gray}: the consolidated Full payload.}
\label{table::mia_decomp}
\setlength{\tabcolsep}{1.4mm}{
\begin{tabular}{l cccc}
    \toprule
    Method & $\mathcal{D}_f$ & $\mathcal{U}$ & $\mathcal{D}_r$ & retained, held out \\
    \midrule
    Original    & 0.996 & 0.996 & 0.996 & 0.997 \\
    Fine-tune   & 0.990 & 0.982 & 0.999 & 0.998 \\
    GA          & 0.994 & 0.994 & 0.991 & 0.996 \\
    \midrule
    Retrain     & \textcolor{blue}{0.722} & \textcolor{blue}{0.493} & \textcolor{blue}{1.000} & \textcolor{blue}{0.938} \\
    \midrule
    \textbf{Core}   & \textbf{\textcolor{red}{0.725}} & \textbf{\textcolor{red}{0.493}} & \textbf{\textcolor{red}{1.000}} & \textbf{\textcolor{red}{0.945}} \\
    \textbf{Full}   & \textcolor{gray}{0.998} & \textcolor{gray}{0.998} & \textcolor{gray}{0.997} & \textcolor{gray}{0.999} \\
    \bottomrule
\end{tabular}}
\end{table}

\begin{table}[htbp]
\centering
\caption{Effect of freezing BN running statistics during Step 2 on Market-1501, under identity-level deletion.
Freezing trades Full capacity for Core accuracy, $-1.9$ and $+0.8$ mAP, and moves the membership attack on $\mathcal{U}$ by $0.006$, so the running statistics are not a leakage channel here.
Default is the deployed setting.}
\label{table::bn_freeze}
\setlength{\tabcolsep}{1.2mm}{
\begin{tabular}{l ccc}
    \toprule
    Setting & Core mAP & Full mAP & MIA$_\text{ent}$ on $\mathcal{U}$ \\
    \midrule
    Default   & 74.34 & 81.99 & 0.493 \\
    BN freeze & 75.12 & 80.07 & 0.487 \\
    \bottomrule
\end{tabular}}
\end{table}

\emph{The low false positive region.} An ROC-AUC averages over every operating point, including ones no adversary would use.
A membership claim is made where the false positive rate is small, so \cref{tab::lowfpr} reports the true positive rate at $1\%$ and $0.1\%$ false positives alongside the AUC.
The size of the non-member pool sets what these columns can resolve, and it is small.
On Market-1501 the attack scores $6473$ members against $3368$ non-members, so $1\%$ false positives means $33$ of them and $0.1\%$ means $3$, a threshold fixed by three samples.
A rate of $0.01\%$ is not attainable with this many non-members and is not reported.
VeRi-776 offers $1678$ non-members and RegDB $2060$, which is why \cref{tab::lowfpr3} carries the $1\%$ column alone across the three benchmarks.
The ordering is the same but the separation is much wider.
At $1\%$ false positives the retrained reference recovers $1.2\%$ of the forgotten members, which is chance, and Core recovers $1.0\%$.
Two other rows come within that range, LOFT at $1.3\%$ and NegGrad+ at its most aggressive rate at $1.7\%$, and neither is forgetting: LOFT still retrieves the deleted identities at $99.6$ mAP, and that NegGrad+ point retrieves at $28.2$ mAP overall.
Every remaining row recovers between $4.0$ and $80.5\%$.
This column is therefore not a ranking on its own, which is the same reason the AUC is not one.
The consolidated Full recovers $38.2\%$, which is the payload being detected as designed.
\cref{tab::lowfpr3} repeats the $1\%$ column on all three benchmarks.
The reading is the same everywhere: the reference recovers $0.012$, $0.004$ and $0.011$ of the members, Core recovers $0.010$, $0.017$ and $0.020$, and the payload recovers $0.382$, $0.205$ and $0.184$.
SSD is the one row that falls below chance on two benchmarks, which is the over-forgetting direction rather than a privacy gain.
LOFT is the sharpest case of a row that reads clean in one place and not in another: $0.013$ on Market-1501, which is chance, against $0.596$ on RegDB, where the same method recovers three in five of the deleted images at a $1\%$ false positive rate.

\begin{table}[htbp]
\centering
\caption{\textbf{Attacks in the low false positive region} on Market-1501 under identity-level deletion, members $\mathcal{U}$ against the query split, deployed seed, four retrained references.
TPR columns are the true positive rate at the stated false positive rate, so $0.010$ at $1\%$ is chance. {\color{red}Red}: Core; {\color{blue}blue}: retrained reference; \textcolor{gray}{gray}: the consolidated Full payload.}
\label{tab::lowfpr}
\setlength{\tabcolsep}{0.3mm}
\begin{tabular}{l ccc cc}
    \toprule
    & \multicolumn{3}{c}{RMIA} & \multicolumn{2}{c}{U-LiRA} \\
    Method & AUC & TPR@1\% & TPR@0.1\% & AUC & TPR@1\% \\
    \midrule
Original & 0.99 & 0.784 & 0.358 & 0.98 & 0.356 \\
Fine-tune & 0.98 & 0.805 & 0.485 & 0.97 & 0.525 \\
GA & 0.98 & 0.744 & 0.381 & 0.97 & 0.327 \\
    \midrule
NegGrad+ & 0.99 & 0.773 & 0.400 & 0.98 & 0.342 \\
SalUn & 0.88 & 0.302 & 0.122 & 0.87 & 0.200 \\
Bad Teacher & 0.86 & 0.311 & 0.137 & 0.85 & 0.156 \\
LoRA & 0.99 & 0.772 & 0.413 & 0.98 & 0.354 \\
SSD & 0.28 & 0.040 & 0.022 & 0.31 & 0.019 \\
SCRUB & 0.99 & 0.717 & 0.356 & 0.97 & 0.282 \\
LoTUS & 0.78 & 0.161 & 0.040 & 0.74 & 0.158 \\
MUNBa & 0.98 & 0.684 & 0.426 & 0.97 & 0.448 \\
LOFT & 0.32 & 0.013 & 0.002 & 0.39 & 0.014 \\
    \midrule
\textcolor{blue}{Retrain} & \textcolor{blue}{0.51} & \textcolor{blue}{0.012} & \textcolor{blue}{0.000} & \textcolor{blue}{0.51} & \textcolor{blue}{0.009} \\
    \midrule
\textbf{\textcolor{red}{MoME-Core}} & \textbf{\textcolor{red}{0.51}} & \textbf{\textcolor{red}{0.010}} & \textbf{\textcolor{red}{0.001}} & \textbf{\textcolor{red}{0.51}} & \textbf{\textcolor{red}{0.010}} \\
\textcolor{gray}{MoME-Full} & \textcolor{gray}{0.94} & \textcolor{gray}{0.382} & \textcolor{gray}{0.097} & \textcolor{gray}{0.85} & \textcolor{gray}{0.071} \\
    \bottomrule
\end{tabular}
\end{table}

\begin{table}[htbp]
\centering
\caption{\textbf{True positive rate at $1\%$ false positives across the three benchmarks}, RMIA on $\mathcal{U}$ against the query split, deployed seed.
Chance is $0.010$.
The reference sits on it everywhere and Core within a factor of two, the consolidated payload is detected everywhere, and the baselines that keep their retrieval are detected on at least one benchmark each. {\color{red}Red}: Core; {\color{blue}blue}: retrained reference; \textcolor{gray}{gray}: the consolidated Full payload.}
\label{tab::lowfpr3}
\setlength{\tabcolsep}{0.9mm}
\begin{tabular}{l ccc}
    \toprule
    Method & Market-1501 & VeRi-776 & RegDB \\
    \midrule
Original & 0.784 & 0.254 & 0.169 \\
Fine-tune & 0.805 & 0.321 & 0.391 \\
GA & 0.744 & 0.088 & 0.347 \\
    \midrule
NegGrad+ & 0.773 & 0.004 & 0.155 \\
SalUn & 0.302 & 0.061 & 0.293 \\
Bad Teacher & 0.311 & 0.023 & 0.348 \\
LoRA & 0.772 & 0.003 & 0.174 \\
SSD & 0.040 & 0.001 & 0.000 \\
SCRUB & 0.717 & 0.002 & 0.038 \\
LoTUS & 0.161 & 0.001 & 0.098 \\
MUNBa & 0.684 & 0.011 & 0.134 \\
LOFT & 0.013 & 0.003 & 0.596 \\
    \midrule
\textcolor{blue}{Retrain} & \textcolor{blue}{0.012} & \textcolor{blue}{0.004} & \textcolor{blue}{0.011} \\
    \midrule
\textbf{\textcolor{red}{MoME-Core}} & \textbf{\textcolor{red}{0.010}} & \textbf{\textcolor{red}{0.017}} & \textbf{\textcolor{red}{0.020}} \\
\textcolor{gray}{MoME-Full} & \textcolor{gray}{0.382} & \textcolor{gray}{0.205} & \textcolor{gray}{0.184} \\
    \bottomrule
\end{tabular}
\end{table}

\emph{Core against the reference family, not against one reference.} A single pair of numbers cannot support indistinguishability, so \cref{tab::bandgap} gives both sides as distributions on all three benchmarks.
Each family is scored by ranking each of its five retrains against the other four, which is the spread a retrain has against its own kind, and Core's five seeds are read against that rather than against one draw from it.
The gap to any baseline that keeps its retrieval is two orders of magnitude larger on the true positive rate.

\emph{Detection power against a planted violation.} The controls above read a model that was trained honestly.
To measure what the statistic can catch, we train models that silently keep a fraction of the requested identities in the core update, so the deletion request is unchanged and only compliance is violated.
\cref{tab::rhopower} scores them against five clean shadows.
A kept identity pins at the floor of the scale, which is $1/12$ here because the target sits outside the family and the mid-rank runs over six models.
Every clean shadow beats the target on its images, while a clean forgotten identity stays uniform and reaches the floor only $2$ to $7\%$ of the time.
The aggregate separates from the null at a $5\%$ violation and beyond, by $2.3$ to $4.5$ standard deviations of the clean family, whose five leave-one-out readings span $0.4908$ to $0.5140$.
The $5\%$ and $10\%$ readings differ by less than half that spread, so their order is not resolved.
It misses the $2\%$ case, where the reading even drifts high because the honest $98\%$ of $\mathcal{U}$ dilutes the leak.
Ranking identities by their mean rank recovers it: the identity-level ROC-AUC is above $0.99$ at every fraction, including the fourteen-identity leak the aggregate cannot see.
In the form an audit would use it, opening as many identities as were kept, that is $13$ of $14$ at $2\%$, $31$ of $34$ at $5\%$, $59$ of $68$ at $10\%$ and $158$ of $169$ at $25\%$ out of the $676$ requested, and at $2\%$ every kept identity falls inside the top fifteen of that ranking.
The same models settle whether anything other than the rank test sees these violations.
Scored with the three membership attacks against the four disjoint references the main tables use, a control trained by the same recipe with the violation set to nothing reads $0.499$ on RMIA and the four violations read $0.531$, $0.542$, $0.562$ and $0.635$.
The calibrated intervals clear the control's from $5\%$ up and overlap it at $2\%$, by $0.005$ on RMIA and $0.001$ on U-LiRA; the uncalibrated entropy attack clears it only from $10\%$.
So the smallest violation any of the three resolves is $5\%$ of the request, the same as the aggregate rank reading and more than twice the fourteen identities the identity-level reading localizes at $0.9995$.

These controls were trained on a $90\%$ deletion split rather than the deployed one, and the five clean shadows are MoME cores rather than from-scratch retrains, which is why their leave-one-out spread of $0.0079$ is an order of magnitude tighter than the $0.085$ a retrain family spans at the deployed split.
The $\sigma$ column therefore measures the power of the test, not the width of the band a deployed audit is read against.

\begin{table}[htbp]
\centering
\caption{\textbf{Detection power of the Forgetting-Rank against planted violations} on Market-1501, five clean shadows, split seed $42$.
A violation of $f$ keeps $f$ of the requested identities in the core update.
The clean leave-one-out readings are $0.4946$, $0.5140$, $0.4999$, $0.4908$ and $0.5006$, a spread of $0.0079$ that sets the $\sigma$ column, which counts standard deviations from the null.
The clean-at-floor column is the fraction of genuinely forgotten identities that also reach it, which is the false positive rate of the identity-level reading.
Bold marks the $2\%$ violation, the one neither the aggregate nor the attacks can see and the identity-level reading still localizes.
The RMIA columns score the same models against the four disjoint references the main tables use, identity-clustered; a control trained by the same recipe with the violation set to nothing reads $0.499$ there, with an interval of $[.482,.517]$.}
\label{tab::rhopower}
\setlength{\tabcolsep}{0.45mm}
\scriptsize
\begin{tabular}{l c c c c c c c}
    \toprule
    & & \multicolumn{4}{c}{Forgetting-Rank} & \multicolumn{2}{c}{RMIA} \\
    \cmidrule(lr){3-6}\cmidrule(lr){7-8}
    Violation & Kept & $\rho_{\mathcal U}$ & $\sigma$ & Identity AUC & Clean at floor & AUC & $95\%$ int. \\
    \midrule
    $2\%$  & 14  & 0.5102 & $+1.3$ & \textbf{0.9995} & 2.0\% & 0.531 & $[.512,.550]$ \\
    $5\%$  & 34  & 0.4793 & $-2.6$ & 0.9973 & 5.5\% & 0.542 & $[.522,.562]$ \\
    $10\%$ & 68  & 0.4820 & $-2.3$ & 0.9921 & 7.2\% & 0.562 & $[.541,.585]$ \\
    $25\%$ & 169 & 0.4645 & $-4.5$ & 0.9940 & 7.3\% & 0.635 & $[.611,.659]$ \\
    \bottomrule
\end{tabular}
\end{table}

\end{document}